\documentclass[runningheads]{llncs}
\usepackage[table]{xcolor}
\usepackage{graphicx}

\usepackage{tikz}
\usepackage{comment}
\usepackage{amsmath,amssymb} %
\usepackage{color}

\usepackage[accsupp]{axessibility}  %

\usepackage{booktabs}
\usepackage[pagebackref,breaklinks,colorlinks=true,linkcolor=RoyalBlue,citecolor=RoyalBlue]{hyperref}

\usepackage[capitalize]{cleveref}

\usepackage[utf8]{inputenc} %
\usepackage[T1]{fontenc}    %
\usepackage{url}            %
\usepackage{amsfonts}       %
\usepackage{nicefrac}       %
\usepackage{microtype}      %
\usepackage{graphicx,color,bm,relsize,enumitem,multirow,epsfig,bbm,adjustbox}
\usepackage{subcaption}

\definecolor{RoyalBlue}{rgb}{0,0,0.8}
\usepackage{wrapfig}
\usepackage{diagbox}
\usepackage{array}
\usepackage{algorithm}
\usepackage{algorithmic}
\newcolumntype{P}[1]{>{\centering\arraybackslash}p{#1}}

\newcommand{\ie}{\textit{i}.\textit{e}., }
\newcommand{\eg}{\textit{e}.\textit{g}., }
\newcommand{\etal}{\textit{et} \textit{al}.}

\newcommand{\ms}[1]{\tiny{$\pm$#1}}
\newcommand{\msf}[1]{\footnotesize{$\pm$#1}}

\begin{document}
\pagestyle{headings}
\mainmatter
\def\ECCVSubNumber{49}  %

\title{OpenCoS: Contrastive Semi-supervised Learning for Handling Open-set Unlabeled Data} %
\titlerunning{Contrastive Semi-supervised Learning for Handling Open-set Unlabeled Data}
\author{Jongjin Park\thanks{Equal contribution}\and
Sukmin Yun$^\star$\and
Jongheon Jeong\and
Jinwoo Shin
}
\authorrunning{J. Park et al.}
\institute{Korea Advanced Institute of Science and Technology (KAIST) \\
\email{\{jongjin.park, sukmin.yun, jongheonj, jinwoos\}@kaist.ac.kr}}
\maketitle

\begin{abstract}
Semi-supervised learning (SSL) has been a powerful strategy to incorporate few labels in learning better representations. 
In this paper, we focus on a practical scenario that one aims to apply SSL when unlabeled data may contain \emph{out-of-class} samples -
those that cannot have one-hot encoded labels from a closed-set of classes in label data, \ie the unlabeled data is an \emph{open-set}. Specifically, we introduce \emph{OpenCoS}, a simple framework for handling this realistic
semi-supervised learning scenario based upon a recent framework of self-supervised visual representation learning. 
We first observe that the out-of-class samples in the open-set unlabeled dataset can be identified effectively via
self-supervised contrastive learning.
Then, OpenCoS utilizes this information to overcome the failure modes in the existing state-of-the-art semi-supervised methods, 
by utilizing one-hot pseudo-labels and soft-labels for the identified in- and out-of-class unlabeled data, respectively.
Our extensive experimental results show the effectiveness of OpenCoS under the presence of out-of-class samples, fixing up the state-of-the-art semi-supervised methods to be suitable for diverse scenarios involving open-set unlabeled data.
\end{abstract}

\section{Introduction}

Despite the recent success of deep neural networks with large-scale labeled data, many real-world scenarios suffer from expensive data acquisition and labeling costs.
This has motivated 
the community
to develop \emph{semi-supervised learning} (SSL;~\cite{grandvalet2005semi,chapelle2009semi}),
\ie by further incorporating unlabeled data for training.
Indeed, recent SSL works~\cite{berthelot2019mixmatch,berthelot2020remixmatch,sohn2020fixmatch,chen2020big} demonstrate promising results on several benchmark datasets, as they could even approach the performance of fully supervised learning using only a small number of labels, \eg 93.73\% accuracy on CIFAR-10 with 250 labeled data~\cite{berthelot2020remixmatch}.

\begin{figure}[t]
  \centering
  \includegraphics[width=0.475\textwidth]{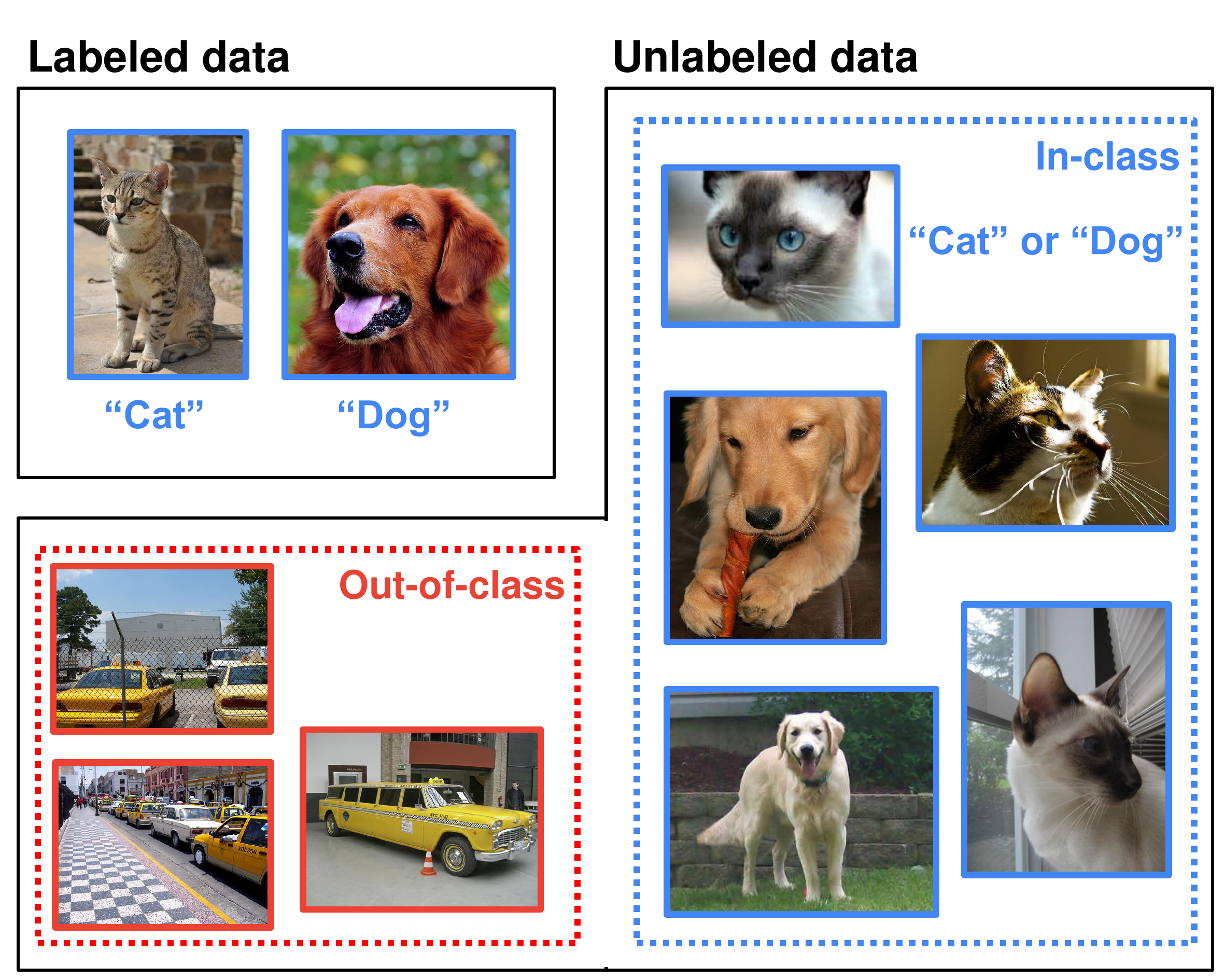}
  \caption{Illustration of 
  an open-set unlabeled data
  under class-distribution mismatch
  in semi-supervised learning, 
  \ie unlabeled data may contain unknown out-of-class samples. 
  }
  \vspace{-0.2in}
 \label{figure_1}
\end{figure}

However, SSL methods often fail to generalize when there is a mismatch between the class-distributions of labeled and unlabeled data~\cite{oliver2018realistic,chensemi,guo2020safe,saito2021openmatch}, \ie when the unlabeled data contains \emph{out-of-class} samples, whose ground-truth labels are not contained in the labeled dataset
(as illustrated in \cref{figure_1}). 
In this scenario, various label-guessing techniques used in the existing SSL methods may label those out-of-class samples incorrectly, 
which in turn significantly
harms the overall training through their inner-process of entropy minimization~\cite{grandvalet2005semi,lee2013pseudo} or consistency regularization~\cite{xie2019unsupervised,sohn2020fixmatch}.
This problem may largely hinder the existing SSL methods from being used in practice, considering the \emph{open-set} nature of unlabeled data collected in the wild \cite{bendale2016towards}.

\begin{figure*}[t]
  \centering
  \includegraphics[width=0.9\textwidth]{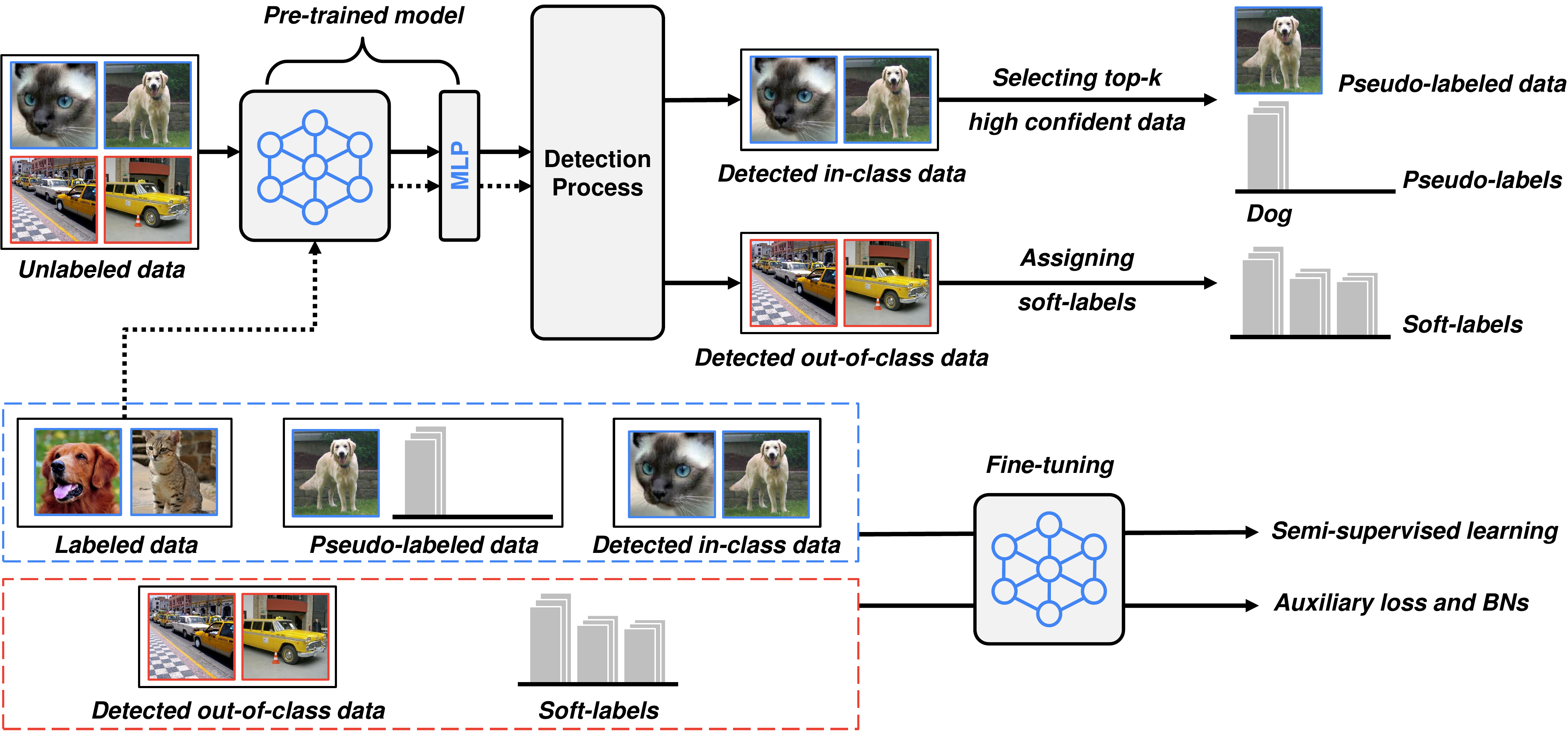} 
  \caption{Overview of our proposed framework, \emph{OpenCoS}. 
  First, our method detects in- and out-of-class samples
  {based on} contrastive representation. 
  {Then, OpenCoS chooses \emph{top-k} confident in-class samples to assign one-hot encoded pseudo-labels, and integrate them into the original labeled dataset for semi-supervised learning.}
  On the other hand, the out-of-class samples detected by OpenCoS are further utilized via an auxiliary loss and batch normalization layers with soft-labels
  generated from the representation.
  }
  \label{fig:main}
  \vspace{-0.1in}
\end{figure*}

\vspace{0.03in}
\noindent\textbf{Contribution.}
In this paper, we focus on a realistic SSL scenario, where unlabeled data may contain some unknown \emph{out-of-class samples}, \ie there is a class distribution mismatch between labeled and unlabeled data~\cite{oliver2018realistic}.
Compared to prior approaches that have bypassed this problem by simply filtering out them \cite{nair2019realmix,chensemi,saito2021openmatch}, %
the unique characteristic in our approach is to further leverage the information in out-of-class samples by assigning \emph{soft-labels} to them:
some of them may still contain useful features for the in-classes.

Our first finding is that a recent technique of \emph{self-supervised representation learning}~\cite{wu2018unsupervised,he2019momentum,chen2020simple,caron2021emerging} can play a key role for our goal. More specifically, we show that a pre-trained representation via self-supervised contrastive learning,
namely {SimCLR \cite{chen2020simple}},
on both labeled and unlabeled data enables us to design (a) an effective score for detecting out-of-class samples in unlabeled data, and (b) a systematic way to 
assign labels to the detected in- and out-of-class samples separately 
by modeling \emph{posterior predictive distributions} using the labeled samples.
Finally, we found (c) auxiliary batch normalization layers~\cite{xie2019adversarial} could 
further help to mitigate the class-distribution mismatch via decoupling batch normalization layers.
We propose a generic SSL framework, coined \emph{OpenCoS}, based on the aforementioned techniques 
for handling open-set unlabeled data, which can be integrated with any existing SSL methods,
\eg ReMixMatch~\cite{berthelot2020remixmatch} and \mbox{FixMatch}~\cite{sohn2020fixmatch}.

We verify the effectiveness of the proposed method on a wide range of SSL benchmarks based on
CIFAR-10, CIFAR-100~\cite{krizhevsky2009learning}, and ImageNet~\cite{deng2009imagenet} datasets, 
assuming the presence of various out-of-class data, \eg SVHN~\cite{netzer2011reading} and
Tiny-ImageNet datasets.
Our experimental results demonstrate that OpenCoS greatly improves existing state-of-the-art SSL methods \cite{berthelot2020remixmatch,sohn2020fixmatch}.
We also compare our method to other recent works~\cite{nair2019realmix,chensemi,guo2020safe,saito2021openmatch} addressing the same class distribution mismatch problem in SSL, and again confirms the effectiveness of our framework,
\eg we achieve an accuracy of {69.38\% with 40 labels (just 4 labels per class) on CIFAR-10 with \mbox{TinyImageNet} as out-of-class, while the recent baseline, OpenMatch~\cite{saito2021openmatch} does 62.71\%.} 

Overall, our work highlights the benefit of unsupervised representations in semi-supervised learning, {which was also explored by~\cite{chen2020big} under a different perspective.}
We newly found that such a \emph{label-free} representation turns out to enhance model generalization due to its (i) robustness on the novel, out-of-class samples,
(ii) successful performance on identifying out-of-class samples,
and (iii) label-efficient transferability (of high confident samples) to the downstream task.

\nocite{miyato2018virtual}
\nocite{zhai2019s4l}
\nocite{raina2007self}

\section{Preliminaries}
\subsection{Semi-supervised learning}\label{subsection:semiintro}
The goal of \emph{semi-supervised learning} {for classification} is to train a classifier $f:\mathcal{X}\rightarrow \mathcal{Y}$ from a \emph{labeled dataset} $\mathcal{D}_l=\{x_{l}^{(i)}, y_l^{(i)}\}_{i=1}^{N_l}$ where each label $y_l$ is from a set of classes $\mathcal{Y}:=\{1, \cdots, C\}$, and an \emph{unlabeled dataset} $\mathcal{D}_u=\{x_{u}^{(i)}\}_{i=1}^{N_u}$ where each $y_u$ exists but is assumed to be unknown.
In an attempt to leverage the extra information in $\mathcal{D}_u$, a number of techniques have been proposed, \eg entropy minimization~\cite{grandvalet2005semi,lee2013pseudo} and consistency regularization~\cite{sajjadi2016regularization,miyato2018virtual,xie2019unsupervised,sohn2020fixmatch}. 
In general, recent approaches in semi-supervised learning can be distinguished 
by the prior they adopt for {the representation of} unlabeled data:
for example, the consistency regularization technique attempt to minimize the cross-entropy {loss} between any two predictions of different augmentations $t_1(x_u)$ and $t_2(x_u)$ from a given unlabeled sample $x_u$, jointly with the standard training for a labeled sample $(x_l, y_l)$:
\begin{equation}\label{eq:semi_loss}
    \mathcal{L}_{\text{SSL}}(x_l, x_u) := \mathbb{H}(y_l, f(x_{l}))
    +\beta\cdot\mathbb{H}(f(t_1(x_u)), f(t_2(x_u))),
\end{equation}
where $\mathbb{H}$ is a standard cross-entropy loss for labeled data, and $\beta$ is a hyperparameter.
Recently, several ``holistic'' approaches of various techniques \cite{zhang2017mixup,cubuk2019randaugment} have shown remarkable performance in practice, \eg MixMatch \cite{berthelot2019mixmatch}, ReMixMatch \cite{berthelot2020remixmatch}, and FixMatch \cite{sohn2020fixmatch}, which we mainly consider in this paper.
We note that our scheme can be integrated with any {recent} semi-supervised learning methods.
\subsection{Contrastive representation learning}\label{subsection:contrastive}
Contrastive learning \cite{oord2018representation,hnaff2019cpcv2,he2019momentum,chen2020simple} defines an unsupervised task for an encoder $f_e: \mathcal{X} \rightarrow \mathbb{R}^{d_e}$ from a set of samples $\{x_i\}$:
assume that a ``query'' sample $x_q$ is given and there is a positive ``key'' $x_{+}\in \{x_i\}$ that $x_q$ matches. Then the \emph{contrastive loss} is defined to let $f_e$ extract the necessary information to identify $x_{+}$ from $x_q$ as follows:
\begin{equation}\label{eq:contrastive}
     \mathcal{L}^{\text{con}}(f_e, x_q, x_{+}; \{x_i\})
     := -\log\frac{\exp(h(f_e(x_q), f_e(x_+))/\tau)}{\sum_{i} \exp(h(f_e(x_q), f_e(x_i))/\tau)},
\end{equation}
where $h(\cdot, \cdot)$ is a pre-defined similarity score, and $\tau$ is a temperature hyperparameter. 
In this paper, we primarily focus on \emph{SimCLR} \cite{chen2020simple}, a particular form of contrastive learning: for a given $\{x_i\}_{i=1}^N$, SimCLR first samples two separate data augmentation operations from a pre-defined family $\mathcal{T}$, namely $t_1, t_2\sim \mathcal{T}$, and matches $(\tilde{x}_i, \tilde{x}_{i+N}):=(t_1(x_i), t_2(x_i))$ as a query-key pair interchangeably. The actual loss is then defined as follows:
\begin{align} \label{eq:simclr}
     \mathcal{L}^{\scriptscriptstyle \text{SimCLR}}(f_e; \{x_i\}_{i=1}^N)
     := \frac{1}{2N}\sum_{q=1}^{2N} \mathcal{L}^{\text{con}}(f_e, \tilde{x}_q, \tilde{x}_{(q + N) \bmod 2N}; \{\tilde{x}_i\}_{i=1}^{2N} \setminus \{\tilde{x}_q\}),
\end{align} \vspace{-0.1in}
\begin{align} \label{eq:simclr_h}
     h^{\scriptscriptstyle \text{SimCLR}}(v_1, v_2) := \mathrm{CosineSimilarity}(g(v_1), g(v_2))
     = \frac{g(v_1) \cdot g(v_2)}{||g(v_1)||_2||g(v_2)||_2},
\end{align}
where $g: \mathbb{R}^{d_e} \rightarrow \mathbb{R}^{d_p}$ is a 2-layer neural network called \emph{projection header}.
In other words, the SimCLR loss defines a task to identify a ``semantically equivalent'' sample to $x_q$ up to the set of data augmentations $\mathcal{T}$.

\section{OpenCoS: a framework for open-set SSL} 

We consider semi-supervised classification problems involving $C$ classes.
In addition to the standard assumption of semi-supervised learning (SSL),
we assume that the unlabeled dataset $\mathcal{D}_u$ is \emph{open-set}, \ie the hidden labels $y_u$ of $x_u$ may not be in $\mathcal{Y}:=\{1, \cdots, C\}$. 
In this scenario, existing semi-supervised learning techniques may degrade the classification performance, possibly due to 
incorrect label-guessing procedure for those out-of-class samples. 
In this respect, we introduce \emph{OpenCoS}, 
{a generic method for detecting and labeling in- and out-of-class unlabeled samples in semi-supervised learning.}
Overall, our key intuition is to utilize the unsupervised representation from \emph{contrastive learning} \cite{wu2018unsupervised,he2019momentum,chen2020simple,caron2021emerging}
to leverage {open-set unlabeled data} in an appropriate manner. 
We present a brief overview of our method in \cref{subsection:overview}, 
and describe 
how our approach can handle {a realistic SSL scenario} in \cref{subsection:stage1} and \cref{subsection:stage2}.

\subsection{Overview of OpenCoS}\label{subsection:overview}

Recall that our goal is to train a classifier $f:\mathcal{X}\rightarrow \mathcal{Y}$ from a labeled dataset $\mathcal{D}_l$ and an \emph{open-set} unlabeled dataset $\mathcal{D}_u$.
Overall, OpenCoS aims to overcome the presence of out-of-class samples in $\mathcal{D}_u$ through the following procedure:
\begin{enumerate}
    \item \textbf{Pre-training via contrastive learning. } OpenCoS first learns an unsupervised representation of $f$ via SimCLR\footnote{
    Nevertheless, our framework is not restricted to a single method of SimCLR: 
    {
    \eg we also show that OpenCoS can be applicable with DINO~\cite{caron2021emerging}, another recent self-supervised learning scheme (see \cref{abl:total}).}}
    \cite{chen2020simple}, using both $\mathcal{D}_l$ and $\mathcal{D}_u$ without labels. 
    More specifically, we learn the penultimate features of $f$, denoted by $f_e$, by minimizing the contrastive loss defined in (\ref{eq:simclr}).
    We also introduce a projection header $g$ (\ref{eq:simclr_h}), which is a 2-layer MLP as per \cite{chen2020simple}.
    \item \textbf{Detecting in- and out-of-class samples. } From a learned representation of $f_e$ and $g$, 
    {OpenCoS identifies \emph{out-of-class} dataset $\mathcal{D}_u^{\text{out}}$ from the given unlabeled dataset $\mathcal{D}_u$.
    Detected out-of-class dataset by our method is denoted as $\widetilde{\mathcal{D}}_u^{\text{out}}$.}
    This detection process is based on the similarity score between $\mathcal{D}_l$ and $\mathcal{D}_u$ in the representation space of $f_e$ and $g$. 
    {Furthermore, \mbox{OpenCoS} constructs a pseudo-labeled dataset $\mathcal{D}_{l}^{\text{pseudo}}$ to enlarge $\mathcal{D}_l$, 
    by assigning one-hot labels to \emph{top-k} confident samples in
    {detected in-class dataset $\widetilde{\mathcal{D}}_u^{\text{in}}:=\mathcal{D}_u \backslash \widetilde{\mathcal{D}}_u^{\text{out}}$} %
    (see \cref{subsection:stage1} for more details).}
    \item {\textbf{Semi-supervised learning with auxiliary loss and batch normalization.}} 
    Now, one can use 
    any existing semi-supervised learning scheme, \eg ReMixMatch \cite{berthelot2020remixmatch}, to train $f$ using labeled dataset {$\mathcal{D}_{l}\cup\mathcal{D}_{l}^{\text{pseudo}}$} and unlabeled dataset $\widetilde{\mathcal{D}}_u^{\text{in}}$. 
    In addition, OpenCoS minimizes an \emph{auxiliary loss} that assigns a soft-label to each sample in $\widetilde{\mathcal{D}}_u^{\text{out}}$, which is also based on the representation of $f_e$ and $g$ (see  \cref{subsection:stage2}).
    Furthermore, we found maintaining auxiliary batch normalization layers \cite{xie2019adversarial} for $\widetilde{\mathcal{D}}_u^{\text{out}}$ is beneficial to our loss
    as they mitigate the distribution mismatch arisen from $\widetilde{\mathcal{D}}_u^{\text{out}}$.
\end{enumerate}
Putting it all together, OpenCoS provides an effective and systematic way to {utilize open-set unlabeled data}
for semi-supervised learning. 
{\cref{fig:main} and \cref{alg:opencos} describe the overall training scheme of OpenCoS.}

\subsection{Detection criterion}\label{subsection:stage1}
\vspace{-0.03in}
For a given labeled dataset $\mathcal{D}_l$ and an open-set unlabeled dataset $\mathcal{D}_u$, 
we aim to detect a subset of the unlabeled training data $\mathcal{D}_u^{\text{out}} \subseteq \mathcal{D}_u$ whose elements are out-of-class, \ie $y_u\notin \mathcal{Y}$. 
{A popular way to handle this task is to use a confidence-calibrated 
classifier trained by $\mathcal{D}_l$ to detect $\mathcal{D}_u^{\text{out}}$ \cite{hendrycks2016baseline,liang2017enhancing,lee2017training,lee2018simple,hendrycks2018deep,hendrycks2019using,bergman2020classification,tack2020csi}.}
However, training such a good classifier is infeasible in our setup due to 
the small size of $\mathcal{D}_l$, \ie lack of labeled data.
Instead, we first perform unsupervised contrastive learning (\ie SimCLR)
using both $\mathcal{D}_l$ and
the open-set unlabeled dataset $\mathcal{D}_u$. %
Then, OpenCoS further
utilizes the labeled dataset $\mathcal{D}_l$ to estimate the class-wise distributions of (pre-trained) embeddings,
and uses them to define a detection score for $\mathcal{D}_u$. 

Formally, we assume that an encoder $f_e: \mathcal{X}\rightarrow \mathbb{R}^{d_e}$ and a projection header $g: \mathbb{R}^{d_e} \rightarrow \mathbb{R}^{d_p}$ pre-trained via SimCLR on $\mathcal{D}_l \cup \mathcal{D}_u$. 
Motivated by the similarity metric used in the pre-training objective of SimCLR (\ref{eq:simclr_h}), we propose a simple yet effective detection score $s(x_u)$
for unlabeled input $x_u$ based on the cosine similarity between $x_u$ and \emph{class-wise prototypical representations} $\{v_c\}_{c=1}^C$ obtained from 
$\mathcal{D}_l$.
Namely, we first define a \emph{class-wise similarity score}
$\mathrm{sim}_c(x_u)$ for each class $c$ as follows:
\vspace{-0.03in}
\begin{align}
    v_c\left(\mathcal{D}_l; f_e, g\right) &:= \frac{1}{N_l^{c}} 
    \sum_i \mathbf{1}_{y_l^{(i)}=c} \cdot g(f_e(x_l^{(i)})), \ \ \text{and} \label{eq:prototype}
\end{align}
\vspace{-0.22in}
\begin{equation}
    \text{sim}_c\left(x_u; \mathcal{D}_l, f_e, g\right) := \mathrm{CosineSimilarity}(g(f_e(x_u)), v_c), %
    \label{eq:similarity score}
\end{equation}
where $N_l^c := |\{(x_l^{(i)}, y_l^{(i)})|y_l^{(i)}=c\}|$ is the sample size of class $c$ in $\mathcal{D}_l$. 
Then, our detection score $s(x_u)$ is defined by the maximal similarity score between $x_u$ and the prototypes $\{v_c\}_{c=1}^C$: 
\vspace{-0.03in}
\begin{equation}
    s(x_u) := \max_{c= 1,\cdots, C}\text{sim}_c\left(x_u\right). \label{eq:confidence score}
\end{equation}
In practice, we use a pre-defined threshold $t$ for detecting out-of-class samples in $\mathcal{D}_u$, \ie we detect a given sample $x_u$ as out-of-class if $s(x_u) < t$. In our experiments, we found an empirical value of $t := \mu_l - 2\sigma_l$ performs well across all the datasets tested, where $\mu_l$ and $\sigma_l$ are mean and standard deviation computed over $\{s(x_l^{(i)})\}_{i=1}^{N_l}$, respectively, 
although more tuning of $t$ could further improve the performance.\footnote{The detection performances with various choices of $t$ are provided in Section~\ref{abl:ths} of the supplementary material.}
Then, we perform such detection process to split the open-set unlabeled dataset $\mathcal{D}_u$ into two datasets $\widetilde{\mathcal{D}}_u^{\text{in}}$ and
$\widetilde{\mathcal{D}}_u^{\text{out}}$, where
$\widetilde{\mathcal{D}}_u^{\text{in}}$ and 
$\widetilde{\mathcal{D}}_u^{\text{out}}=\mathcal{D}_u \backslash \widetilde{\mathcal{D}}_u^{\text{in}}$ are unlabeled datasets
detected to be \emph{in-class} and \emph{out-of-class}, respectively.

\begin{algorithm}[t]
\caption{OpenCoS: A general framework for open-set semi-supervised learning (SSL).}
\label{alg:opencos}
\begin{algorithmic}
\STATE {\bf Input:} Classifier $f$, encoder $f_e$, projection header $g$, labeled dataset $\mathcal{D}_l$, open-set unlabeled dataset $\mathcal{D}_u$. 
\vspace{0.05in}
\hrule
\vspace{0.05in}
\STATE Pre-train $f_e$ and $g$ via contrastive learning using $\mathcal{D}_l\cup\mathcal{D}_u$
\STATE Detect out-of-class samples using threshold $t$ as
\STATE \quad\textcolor{gray}{// $s(\cdot)$ defined in (\ref{eq:confidence score})}
\STATE \quad $\widetilde{\mathcal{D}}_u^{\text{in}}\leftarrow \{x_u\in\mathcal{D}_u|s(x_u; f_e,g)>t\}$
\STATE \quad $\widetilde{\mathcal{D}}_u^{\text{out}}\phantom{.}\leftarrow\mathcal{D}_u \backslash \widetilde{\mathcal{D}}_u^{\text{in}}$ 
\STATE Construct a pseudo-labeled dataset $\mathcal{D}_{l}^{\text{pseudo}}$
\STATE Update $\mathcal{D}_l\leftarrow\mathcal{D}_l\cup\mathcal{D}_{l}^{\text{pseudo}}$
\FOR{each sample $x_{l} \in \mathcal{D}_l$, $x_u^{\text{in}} \in \widetilde{\mathcal{D}}_u^{\text{in}}$, and $x_u^{\text{out}} \in \widetilde{\mathcal{D}}_u^{\text{out}}$}
    \STATE \textcolor{gray}{// $q(\cdot)$ defined in (\ref{eq:similarity})}
    \STATE $\mathcal{L}_{\text{OpenCoS}} \leftarrow \mathcal{L}_{\text{SSL}}(x_{l}, x_u^{\text{in}}; f)
    +\lambda\cdot\mathbb{H}(q(x_u^{\text{out}}), f(x_u^{\text{out}}))$ 
    \STATE Update parameters of $f$ by computing the gradients of \phantom{...} the proposed loss $\mathcal{L}_{\text{OpenCoS}}$
\ENDFOR
\end{algorithmic}
\end{algorithm}

Furthermore, to overcome the regime of few labels in semi-supervised learning, we enlarge the labeled dataset $\mathcal{D}_{l}$ using a subset of $\widetilde{\mathcal{D}}_u^{\text{in}}$ with their  one-hot encoded pseudo-labels, \ie construct an additional (pseudo-) labeled dataset $\mathcal{D}_{l}^{\text{pseudo}}$ to utilize it for semi-supervised learning.
To be specific, we choose \emph{top-k} high confident samples in $\widetilde{\mathcal{D}}_u^{\text{in}}$, under a classifier built upon the pre-trained encoder $f_e$ with the linear evaluation protocol \cite{zhang2016colorful,chen2020simple}.

\subsection{Auxiliary loss and batch normalization}\label{subsection:stage2}
Now, one can train a classifier $f$ on the labeled dataset $\mathcal{D}_{l}\cup\mathcal{D}_{l}^{\text{pseudo}}$ and 
unlabeled dataset 
$\widetilde{\mathcal{D}}_u^{\text{in}}$ under any existing semi-supervised learning method \cite{berthelot2019mixmatch,berthelot2020remixmatch,sohn2020fixmatch}.
In addition, we propose to further utilize $\widetilde{\mathcal{D}}_u^{\text{out}}$ via an auxiliary loss that assigns a soft-label to each $x_u^{\text{out}} \in \mathcal{D}_u^{\text{out}}$. More specifically, for any semi-supervised learning objective $\mathcal{L}_{\text{SSL}}(x_{l}, x_u^{\text{in}}; f)$, we consider the following modified loss to optimize:
\begin{equation}
    \mathcal{L}_{\text{OpenCoS}} = \mathcal{L}_{\text{SSL}}(x_{l}, x_u^{\text{in}}; f)
    +\lambda\cdot\mathbb{H}(q(x_u^{\text{out}}), f(x_u^{\text{out}})),\label{eq:loss}
\end{equation}
where $x_{l} \in \mathcal{D}_{l}\cup\mathcal{D}_{l}^{\text{pseudo}}$,\footnote{If $\mathcal{D}_{l}\cup\mathcal{D}_{l}^{\text{pseudo}}$ is class-imbalanced, we apply oversampling method~\cite{japkowicz2000class} for balancing class distributions.} $\mathbb{H}$ denotes the cross-entropy loss, $\lambda$ is a hyperparameter, and $q(x_u^{\text{out}})$ defines a specific assignment of distribution over $\mathcal{Y}$ for $x_u^{\text{out}}$.
Here, we assign $q(x_u^{\text{out}})$ based on the class-wise similarity scores $\mathrm{sim}_c(x_u^{\text{out}})$ defined in (\ref{eq:similarity score}), again utilizing the contrastive representation $f_e$ and $g$:
\begin{equation}
    q_c(x_u) := \frac{\exp{(\text{sim}_c(x_u; f_e, g)/\tau)}}{\sum_{i}{\exp{(\text{sim}_i(x_u; f_e, g)/\tau)}}},
    \label{eq:similarity}
\end{equation}
where $\tau$ is a temperature hyperparameter.

At first glance, assigning a label of $\mathcal{Y}$ to $x_u^{\text{out}}$ may seem counter-intuitive, as the true label of $x_u^{\text{out}}$ is not in $\mathcal{Y}$ by definition. 
However, even when the out-of-class samples cannot be represented as one-hot labels, 
one can still model their \emph{posterior predictive distributions} over in-classes as a linear combination (\ie soft-label) of $\mathcal{Y}$:
for instance, although ``cat'' images are out-of-class for CIFAR-100, 
still there are some classes in CIFAR-100 that is \emph{semantically similar} to ``cat'', \eg
``leopard'', ``lion'', or ``tiger''. Here, we hypothesize that assigning a soft-label of in-classes is beneficial for such semantically similar classes.
Even if out-of-classes are totally different from in-classes, one can assign the uniform labels to ignore them.
We empirically found that such soft-labels based on the contrastive representation
offer an effective way to utilize out-of-class samples, while they are known to significantly harm in the vanilla semi-supervised learning schemes~\cite{oliver2018realistic,chensemi,guo2020safe}.

\vspace{0.03in}
\noindent\textbf{Auxiliary batch normalization.} %
Finally, we suggest to handle a \emph{data-distribution shift} originated from the class-distribution mismatch~\cite{oliver2018realistic}, 
\ie $\mathcal{D}_l$ and $\mathcal{D}_u^{\text{out}}$ are drawn from the different underlying distribution. 
This may degrade the in-class classification performance as the auxiliary loss utilizes out-of-class samples.
To handle the issue, we use additional batch normalization layers (BN) \cite{ioffe2015batch} for training samples in $\widetilde{\mathcal{D}}_u^{\text{out}}$ to disentangle those two distributions.
In our experiments, we empirically observe such \emph{auxiliary BNs} are beneficial when using 
out-of-class samples via the auxiliary loss.
Auxiliary BNs also have been studied in adversarial learning literature~\cite{xie2019adversarial}:
decoupling BNs improves the performance of adversarial training by handling a distribution mismatch between clean and adversarial samples.
In this paper, we found that a similar strategy 
can improve model performance in realistic semi-supervised learning.

\section{Experiments}\label{sec:exp}
In this section, we {verify} the effectiveness of our method over a wide range of semi-supervised learning (SSL) benchmarks in the presence of various out-of-class data. 
The full details on experimental setups can be found in the supplementary material.

\begin{figure}[t]
  \centering
  \includegraphics[width=0.55\textwidth]{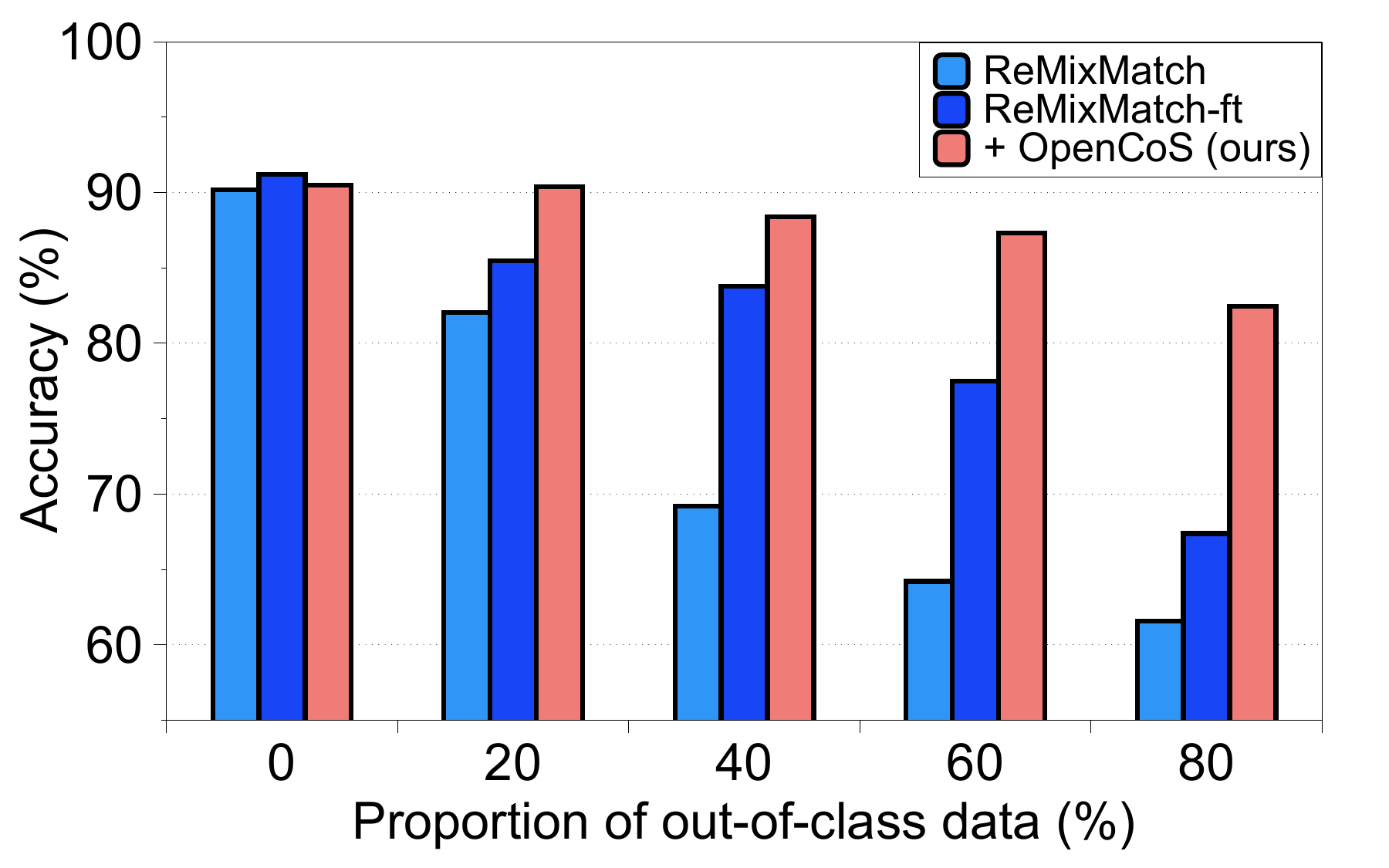}\label{fig:proportion
  }
  \vspace{-0.1in}
  \caption{Comparison of median test accuracy under varying proportions of out-of-class samples on the CIFAR-10 + TinyImageNet benchmark
  with 25 labels per class.
  }
  \label{fig:2}
\end{figure}

\noindent\textbf{Datasets.}~We perform experiments on image classification tasks for several benchmarks in the literature of SSL \cite{berthelot2020remixmatch,sohn2020fixmatch}: CIFAR-10, CIFAR-100~\cite{krizhevsky2009learning}, and ImageNet~\cite{deng2009imagenet} datasets.
Specifically, we focus on settings where each dataset is extremely label-scarce: only 4 or 25 labels per class are given during training, while the rest of the training data are assumed to be unlabeled.
To configure realistic SSL scenarios, 
we additionally assume that unlabeled {data} %
contain samples from an external dataset:
for example, in the case of CIFAR-10,
we use unlabeled samples from SVHN~\cite{netzer2011reading} or TinyImageNet\footnote{\url{https://tiny-imagenet.herokuapp.com/}} datasets.

\begin{figure}[t]
  \centering
  \begin{subfigure}{0.48\linewidth}
  \centering
    \includegraphics[width=\textwidth]{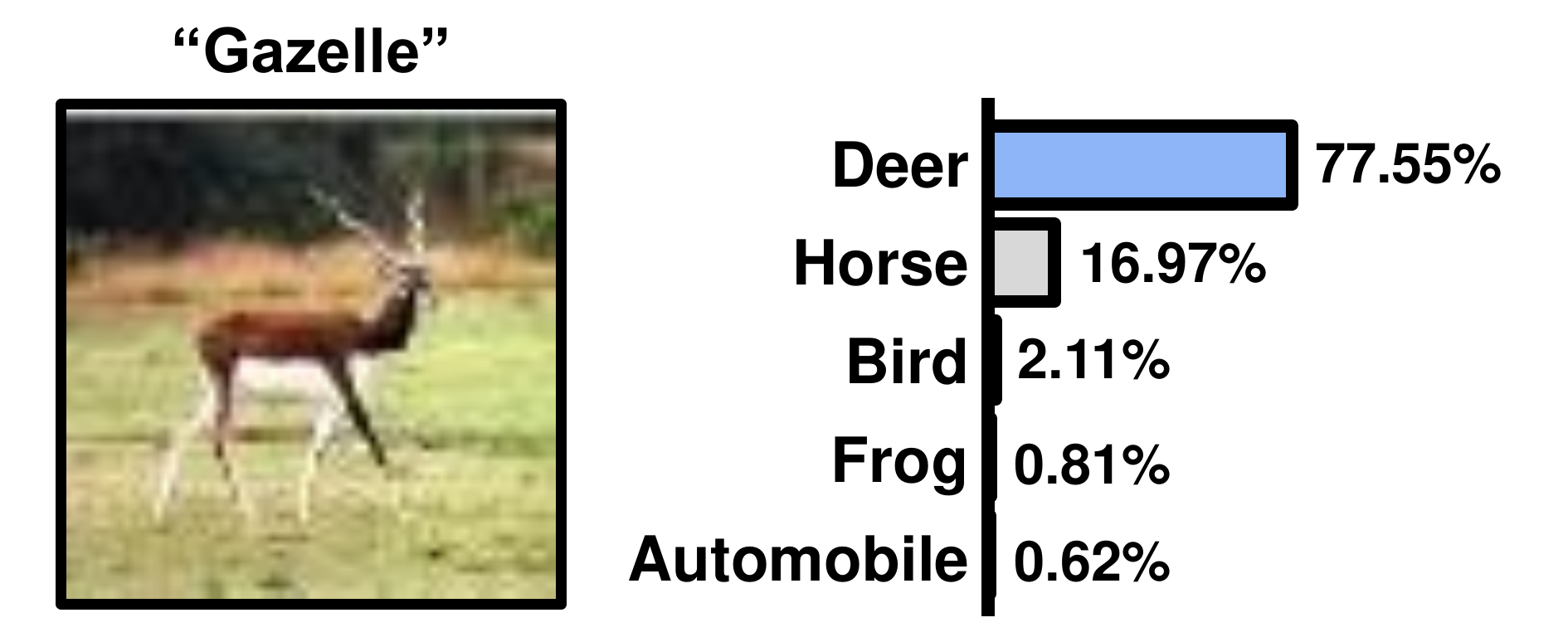}
    \caption{Soft-label assignments of ``Gazelle''.}
    \label{fig:ex2}
  \end{subfigure}
  \begin{subfigure}{0.48\linewidth}
  \centering
    \includegraphics[width=\textwidth]{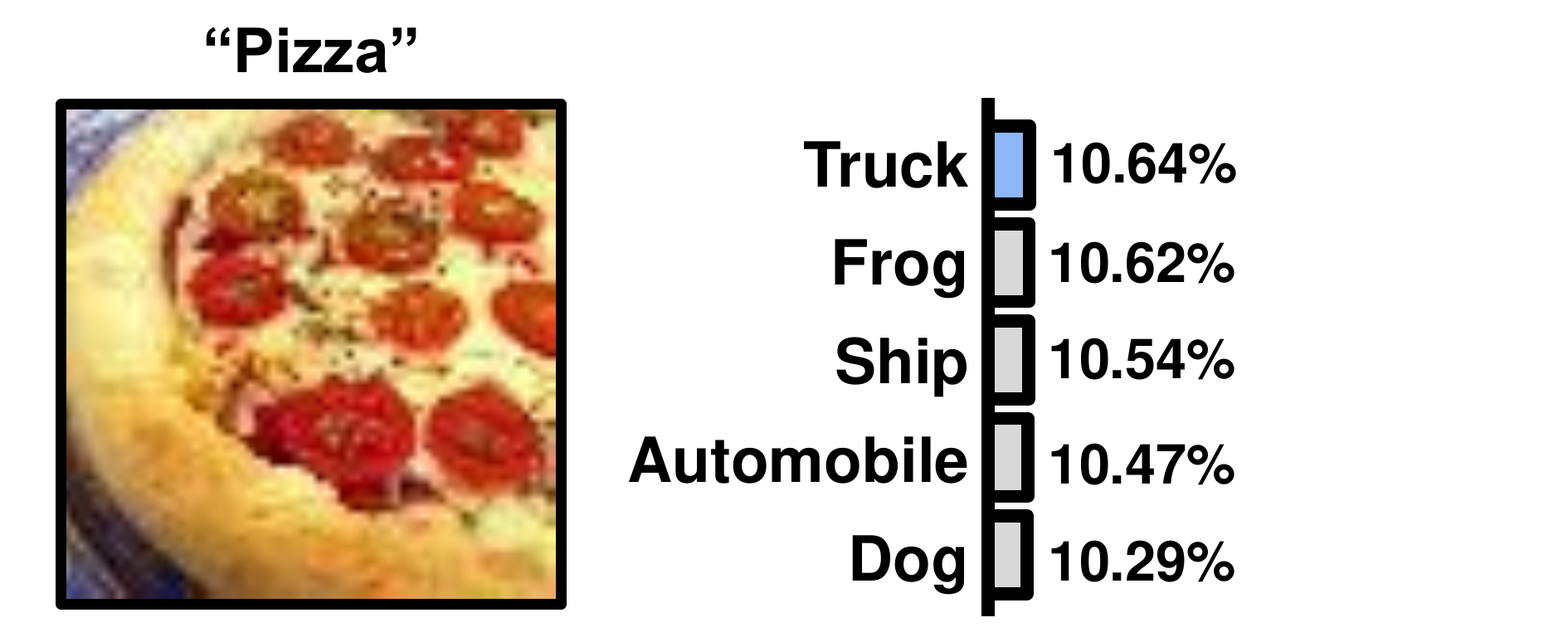}
    \caption{Soft-label assignments of ``Pizza''.}
    \label{fig:ex3}
  \end{subfigure}
  \caption{Illustration of soft-label assignments in the CIFAR-10 + TinyImageNet benchmark. Unlabeled out-of-class samples from
  (a) ``gazelle'' is assigned with soft-labels of $\approx$78\% confidence for ``deer'', and
(b) ``pizza'' is assigned with almost uniform soft-labels ($\approx$10\% of confidence). 
  }
  \label{fig:examples}
\vspace{-0.2in}
\end{figure}

\noindent\textbf{Baselines.}~We evaluate MixMatch~\cite{berthelot2019mixmatch}, ReMixMatch~\cite{berthelot2020remixmatch}, and FixMatch~\cite{sohn2020fixmatch} as baselines in our experimental setup, which are considered to be state-of-the-art methods in conventional SSL.
We also compare our method with 
four prior works applicable to our setting: namely, we consider Uncertainty-Aware Self-Distillation (UASD;~\cite{chensemi}), RealMix~\cite{nair2019realmix}, DS$^3$L~\cite{guo2020safe}, and OpenMatch~\cite{saito2021openmatch},
which propose schemes to detect and filter out out-of-class samples in the unlabeled dataset: 
\eg DS$^3$L re-weights unlabeled samples,
and OpenMatch learns an outlier detector to reduce the effect of such out-of-class samples.
Recall that our method uses \mbox{SimCLR} \cite{chen2020simple} for pre-training.
Unless otherwise noted, we also pre-train the baselines via SimCLR for a fair comparison, denoting those fine-tuned models by ``-ft,'' \eg MixMatch-ft and UASD-ft. 
We confirm that fine-tuned models show comparable or better performance compared to those trained from scratch, as presented in \cref{fig:2}.
Also, we report the performance purely obtainable from (unsupervised) SimCLR: namely, we additionally consider (a) \emph{SimCLR-le}: a SimCLR model with linear evaluation protocol~\cite{zhang2016colorful,chen2020simple}, \ie it additionally learns a linear layer with the labeled dataset, 
(b) \emph{SimCLR-ft}: the whole SimCLR model is fine-tuned with the labeled dataset, 
and (c) \emph{SimCLR-sd}: the self-distilled \mbox{SimCLR} model with a distillation loss of \cite{chen2020big}. 
Somewhat interestingly, these models turn out to be strong baselines in our setups;
they often outperform the state-of-the-art semi-supervised baselines
under large proportions of out-of-class samples (see \cref{tbl:main}).
Finally, we remark that our framework can incorporate any conventional semi-supervised methods for training. 
We denote our method built upon an existing method by \mbox{``+ OpenCoS''}, \eg ReMixMatch-ft + OpenCoS.

\vspace{0.03in}
\noindent\textbf{Training details.}
As suggested by \cite{oliver2018realistic}, we have re-implemented all baseline methods considered, including SimCLR, under the same codebase 
and performed experiments with the same model architecture of ResNet-50~\cite{he2016deep}.\footnote{%
Note that this architecture is larger than Wide-ResNet-28-2
\cite{zagoruyko2016wide}
used in the SSL literature~\cite{oliver2018realistic}. We use ResNet-50 following the standard of SimCLR.} %
We checkpoint per $2^{16}$ training samples and report (a) the median test accuracy of the last 5 checkpoints out of 50 checkpoints in total and (b) the best accuracy among all the checkpoints.
We simply fix $\tau = 0.1$, the temperature hyperparameter in (\ref{eq:similarity}), and $\lambda=0.5$ in (\ref{eq:loss}) in all our experiments.
Finally, we take the top 10\% and 1\% of confident samples for a pseudo-labeled dataset when 4 and 25 labels per class are given, respectively.
The details on model architecture and hyperparameters can be found in the supplementary material.

\begin{table*}[t]
    \centering
    \footnotesize
    \begin{adjustbox}{width=0.97\textwidth}
    \begin{tabular}{lcccccc}
    \toprule
        \multicolumn{2}{l}{In-class} & \multicolumn{1}{c}{CIFAR-Animals} & \multicolumn{2}{c}{CIFAR-10} & \multicolumn{2}{c}{CIFAR-100}  \\
        \cmidrule(l{3pt}r{3pt}){1-2} \cmidrule(l{3pt}r{3pt}){3-3} \cmidrule(l{3pt}r{3pt}){4-5} \cmidrule(l{3pt}r{3pt}){6-7}
        \multicolumn{1}{l}{Out-of-class} &  \multicolumn{1}{l}{Open-SSL}  & \multicolumn{1}{c}{CIFAR-Others} & \multicolumn{1}{c}{SVHN} & \multicolumn{1}{c}{TinyImageNet} &        \multicolumn{1}{c}{SVHN} & \multicolumn{1}{c}{TinyImageNet}\\
        \midrule[0.7pt]
        \multicolumn{7}{c}{\cellcolor{gray! 20}\emph{\# labels per class = 4}}  \\
        \midrule[0.7pt]
        SimCLR-le~\cite{chen2020simple}& - & 65.58\ms{3.51}\phantom{ (xx.xx)} & 56.89\ms{3.19}\phantom{ (xx.xx)} & 58.20\ms{0.88}\phantom{ (xx.xx)} & 22.86\ms{0.17}\phantom{ (xx.xx)} & 27.93\ms{0.67}\phantom{ (xx.xx)} \\
        SimCLR-ft~\cite{chen2020simple} & - & 67.29\ms{2.76} \tiny(68.25) & 42.16\ms{2.50} \tiny(42.67) & 54.26\ms{1.26} \tiny(55.01) & 18.99\ms{0.04} \tiny(19.12) & 29.57\ms{0.33} \tiny(29.57) \\
        SimCLR-sd~\cite{chen2020big} & - & 64.70\ms{3.57} \tiny(67.07) & 56.18\ms{1.63} \tiny(57.22) & 49.88\ms{2.46} \tiny(52.78) & 22.88\ms{0.71} \tiny(23.60) & 26.23\ms{1.32} \tiny(28.26) \\
        \cmidrule(l{3pt}r{3pt}){1-1} \cmidrule(l{3pt}r{3pt}){2-2} \cmidrule(l{3pt}r{3pt}){3-3} \cmidrule(l{3pt}r{3pt}){4-5} \cmidrule(l{3pt}r{3pt}){6-7}
        UASD-ft~\cite{chensemi}       & \checkmark & 43.92\ms{1.94} (52.87) & 42.99\ms{3.05} (44.70) & 50.38\ms{2.78} (51.66) & 19.66\ms{0.44} (19.92) & 25.72\ms{0.69} (26.33) \\
        RealMix-ft~\cite{nair2019realmix}    & \checkmark & 64.42\ms{7.26} (67.99) & 38.22\ms{3.41} (41.55) & 48.28\ms{5.73} (49.78) & 18.48\ms{0.42} (20.04) & 22.14\ms{0.71} (26.51) \\
        DS$^3$L-ft~\cite{guo2020safe}    & \checkmark & 63.98\ms{6.96} (72.20) & 36.81\ms{7.67} (47.32) & 56.32\ms{1.31} (57.58) & 16.35\ms{0.20} (16.97) & 23.95\ms{1.43} (25.06) \\
        OpenMatch-ft~\cite{saito2021openmatch}  & \checkmark & 52.09\ms{5.98} (69.96) & 55.93\ms{2.09} (57.36) & 62.71\ms{1.33} (64.27) & 14.85\ms{1.17} (22.30) & 17.82\ms{2.75} (27.57) \\
        \cmidrule(l{3pt}r{3pt}){1-1} \cmidrule(l{3pt}r{3pt}){2-2} \cmidrule(l{3pt}r{3pt}){3-3} \cmidrule(l{3pt}r{3pt}){4-5} \cmidrule(l{3pt}r{3pt}){6-7}
        MixMatch-ft~\cite{berthelot2019mixmatch}   & - & 44.34\ms{5.13} (65.55) & 23.71\ms{8.65} (38.69) & 38.90\ms{4.24} (46.59) & 13.45\ms{1.23} (16.76) & 23.16\ms{1.85} (26.54) \\
         FixMatch-ft~\cite{sohn2020fixmatch}  & - & 34.94\ms{6.18} (75.83) & 32.70\ms{6.28} (55.58) & 35.99\ms{2.63} (63.35) & 23.56\ms{0.68} (24.24) & 30.70\ms{3.67} (32.52) \\
        \cmidrule(l{3pt}r{3pt}){1-1} \cmidrule(l{3pt}r{3pt}){2-2} \cmidrule(l{3pt}r{3pt}){3-3} \cmidrule(l{3pt}r{3pt}){4-5} \cmidrule(l{3pt}r{3pt}){6-7}
        ReMixMatch-ft~\cite{berthelot2020remixmatch} & - & 47.61\ms{6.51} (64.06) & 24.56\ms{3.99} (47.65) & 28.51\ms{5.87} (55.68) & \phantom{ }9.36\ms{1.97} (21.30) & 22.33\ms{1.10} (29.77) \\
        \textbf{+ OpenCoS (ours)} &  \checkmark & \textbf{81.29\ms{0.93} (81.77)} & \textbf{66.42\ms{7.26} (66.78)}  & \textbf{69.38\ms{3.64} (70.02)} &  \textbf{30.29\ms{2.33} (30.62)}   &     \textbf{36.79\ms{0.97} (37.12)}        \\
        \midrule[0.7pt]
        \multicolumn{7}{c}{\cellcolor{gray! 20}\emph{\# labels per class = 25}}  \\
        \midrule[0.7pt]
        SimCLR-le~\cite{chen2020simple} &-& 80.03\ms{0.73}\phantom{ (xx.xx)} & 70.31\ms{0.14}\phantom{ (xx.xx)} & 71.84\ms{0.10}\phantom{ (xx.xx)} & 37.74\ms{0.42}\phantom{ (xx.xx)} & 43.68\ms{0.26}\phantom{ (xx.xx)} \\
        SimCLR-ft~\cite{chen2020simple} &-& 81.44\ms{0.49} \tiny(81.61) & 64.41\ms{1.37} \tiny(64.65) & 73.05\ms{0.11} \tiny(73.30) & 39.61\ms{0.28} \tiny(39.87) & 49.69\ms{0.30} \tiny(49.96) \\
        SimCLR-sd~\cite{chen2020big}  & - & 80.94\ms{0.93} \tiny(81.82) & 70.49\ms{0.60} \tiny(71.04) & 66.59\ms{0.47} \tiny(69.84) & 38.47\ms{1.08} \tiny(38.98) & 44.09\ms{1.31} \tiny(45.69) \\
        \cmidrule(l{3pt}r{3pt}){1-1} \cmidrule(l{3pt}r{3pt}){2-2} \cmidrule(l{3pt}r{3pt}){3-3} \cmidrule(l{3pt}r{3pt}){4-5} \cmidrule(l{3pt}r{3pt}){6-7}
        UASD-ft~\cite{chensemi}    & \checkmark & 82.17\ms{0.85} (82.50) & 66.70\ms{1.00} (67.43) & 73.97\ms{0.37} (74.54) & 39.51\ms{0.76} (39.65) & 44.58\ms{0.77} (44.90) \\
        RealMix-ft~\cite{nair2019realmix} & \checkmark   & 80.27\ms{2.64} (81.04) & 58.15\ms{5.27} (67.27) & 69.19\ms{2.31} (72.29) & 44.14\ms{1.01} (44.89) & 47.57\ms{1.39} (49.47) \\
        DS$^3$L-ft~\cite{guo2020safe} & \checkmark   & 81.31\ms{0.50} (83.27) & 50.00\ms{8.34} (63.11) & 69.13\ms{2.30} (72.23) & 29.00\ms{0.97} (30.17) & 40.16\ms{0.90} (41.82) \\
        OpenMatch-ft~\cite{saito2021openmatch}   & \checkmark & 86.14\ms{0.30} (86.51) & 74.74\ms{0.98} (75.02) & 77.94\ms{0.76} (78.14) & 29.03\ms{1.26} (39.08) & 32.15\ms{2.62} (47.38) \\
        \cmidrule(l{3pt}r{3pt}){1-1} \cmidrule(l{3pt}r{3pt}){2-2} \cmidrule(l{3pt}r{3pt}){3-3} \cmidrule(l{3pt}r{3pt}){4-5} \cmidrule(l{3pt}r{3pt}){6-7}
        MixMatch-ft~\cite{berthelot2019mixmatch} &- & 83.88\ms{1.60} (84.21) & 17.98\ms{2.60} (54.19) & 69.27\ms{6.59} (75.11) & 38.60\ms{1.86} (43.02) & 50.23\ms{0.89} (51.38) \\
        FixMatch-ft~\cite{sohn2020fixmatch} &- & 69.86\ms{1.92} (84.24) & 68.02\ms{0.68} (71.91) & 70.49\ms{1.15} (77.27) & 41.73\ms{1.29} (42.28) & 45.94\ms{1.03} (49.96) \\
        \cmidrule(l{3pt}r{3pt}){1-1} \cmidrule(l{3pt}r{3pt}){2-2} \cmidrule(l{3pt}r{3pt}){3-3} \cmidrule(l{3pt}r{3pt}){4-5} \cmidrule(l{3pt}r{3pt}){6-7}
        ReMixMatch-ft~\cite{berthelot2020remixmatch} &-& 81.62\ms{1.47} (83.90) & 37.98\ms{3.43} (65.33) & 67.38\ms{7.25} (73.34) & 32.75\ms{0.77} (44.62) & 49.63\ms{1.10} (53.20) \\
        \textbf{+ OpenCoS (ours)} & \checkmark & \textbf{87.15\ms{0.70} (87.72)} & \textbf{78.97\ms{1.13} (79.24)}  & \textbf{82.46\ms{1.33} (82.74)} & \textbf{49.19\ms{1.16} (49.56)}     & \textbf{54.01\ms{1.75} (54.32)}          \\
        \bottomrule
    \end{tabular}
    \end{adjustbox}
    \vspace{0.1in}
    \caption{Comparison of median test accuracy on various benchmark datasets.
    We report the mean and standard deviation over three runs with different random seeds and splits, and also report the mean of the best accuracy in parentheses.
    The best scores are indicated in bold. 
    We denote methods handling unlabeled out-of-class samples (\ie open-set) as ``Open-SSL''.
    }\label{tbl:main}
\end{table*}

\subsection{Effects of out-of-class unlabeled samples}\label{sec:va}
We first evaluate the effect of out-of-class unlabeled samples in semi-supervised learning, on varying proportions to the training dataset. We consider CIFAR-10 and TinyImageNet datasets, and synthetically control the proportion between the two in 50K training samples. For example, 80\% of proportion means the training dataset consists of 40K samples from TinyImageNet, and 10K samples from CIFAR-10. In this experiment, we assume that 25 labels per class are always given in the CIFAR-10 side.
We compare three models on varying proportions of out-of-class: (a) a \mbox{ReMixMatch} model trained from scratch (\mbox{ReMixMatch}), (b) a \mbox{SimCLR} model fine-tuned by \mbox{ReMixMatch} (\mbox{ReMixMatch-ft}), and (c) our OpenCoS model applied to {ReMixMatch-ft} (+ \mbox{OpenCoS}).

\cref{fig:2} demonstrates the results.
Overall, we observe that the performance of ReMixMatch rapidly degrades as the proportion of out-of-class samples increases in unlabeled data. While \mbox{ReMixMatch-ft} significantly mitigates this problem, however, it still fails at a larger proportion: \eg at 80\% of out-of-class, the performance of ReMixMatch-ft falls into that of ReMixMatch.
OpenCoS, 
in contrast, successfully prevents the performance degradation of ReMixMatch-ft,
especially at the regime that out-of-class samples dominate in-class samples.

\begin{table*}[t]
\centering
\centering
\begin{adjustbox}{max width=0.97\textwidth}
{\LARGE
\begin{tabular}{lccccccccccc}
\toprule[1.2pt]
\multicolumn{2}{l}{In-class} & Dog & Reptile & Produce & Bird & Insect & Food & Primate & Aquatic  & Scenery\\
\cmidrule(l{3pt}r{3pt}){1-2} \cmidrule(l{3pt}r{3pt}){3-3} \cmidrule(l{3pt}r{3pt}){4-4} \cmidrule(l{3pt}r{3pt}){5-5} \cmidrule(l{3pt}r{3pt}){6-6} \cmidrule(l{3pt}r{3pt}){7-7} \cmidrule(l{3pt}r{3pt}){8-8} \cmidrule(l{3pt}r{3pt}){9-9}
\cmidrule(l{3pt}r{3pt}){9-9} \cmidrule(l{3pt}r{3pt}){10-10} \cmidrule(l{3pt}r{3pt}){11-11}
{Number of class} & {Open-SSL} & 118 & 36 & 22 & 21 & 20 & 19 & 18 & 13 & 11\\
\cmidrule(l{3pt}r{3pt}){1-1} \cmidrule(l{3pt}r{3pt}){2-2} \cmidrule(l{3pt}r{3pt}){3-11} 
SimCLR-le~\cite{chen2020simple} &-& 43.02\msf{0.56} & 51.76\msf{0.92} & 64.76\msf{0.58} & 75.81\msf{1.01} & 59.90\msf{0.92} & 56.53\msf{0.73} & 53.67\msf{0.69} & 68.41\msf{1.31} & 64.73\msf{0.73} \\
SimCLR-ft~\cite{chen2020simple}&-& 46.72\msf{0.63} & 51.76\msf{1.42} & 65.21\msf{1.05} & 77.37\msf{0.95} & 58.93\msf{1.50} & 54.63\msf{0.79} & 55.29\msf{1.79} & 68.82\msf{2.15} & 62.79\msf{1.15}\\
SimCLR-sd~\cite{chen2020big}&-& 36.12\msf{5.76} & 49.35\msf{1.22} & 61.88\msf{1.38} & 75.71\msf{1.72} & 57.13\msf{2.21} & 57.49\msf{9.99} & 47.70\msf{4.71} & 68.72\msf{0.70} & 66.31\msf{1.03}\\
\cmidrule(l{3pt}r{3pt}){1-1} \cmidrule(l{3pt}r{3pt}){2-2} \cmidrule(l{3pt}r{3pt}){3-3} \cmidrule(l{3pt}r{3pt}){4-4}
\cmidrule(l{3pt}r{3pt}){5-5} \cmidrule(l{3pt}r{3pt}){6-6} \cmidrule(l{3pt}r{3pt}){7-7} \cmidrule(l{3pt}r{3pt}){8-8}
\cmidrule(l{3pt}r{3pt}){9-9} \cmidrule(l{3pt}r{3pt}){10-10} \cmidrule(l{3pt}r{3pt}){11-11}
UASD-ft~\cite{chensemi}&\checkmark& 45.64\msf{0.89} & 53.07\msf{0.73} & 67.09\msf{0.65} & 78.92\msf{0.40} & 61.53\msf{1.56} & 55.90\msf{1.04} & 56.70\msf{0.85} & 70.31\msf{0.85} & 64.36\msf{1.13}\\
RealMix-ft~\cite{nair2019realmix}&\checkmark& 43.55\msf{2.36} & 45.70\msf{0.16} & 56.06\msf{0.65} & 71.94\msf{1.21} & 53.33\msf{1.78} & 48.25\msf{1.30} & 45.89\msf{0.84} & 58.10\msf{1.34} & 60.79\msf{1.36}\\
DS$^3$L-ft~\cite{guo2020safe}&\checkmark& 44.12\msf{1.37} & 52.09\msf{1.47} & 65.39\msf{1.22} & 78.00\msf{0.25} & 59.30\msf{2.33} & 54.32\msf{1.37} & 53.44\msf{0.80} & 70.67\msf{1.23} & 62.67\msf{2.37}\\
OpenMatch-ft~\cite{saito2021openmatch}&\checkmark& 48.85\msf{2.85} & 54.55\msf{1.98} & 67.21\msf{1.37} & 78.60\msf{1.80} & 63.20\msf{1.76} & 55.86\msf{1.85} & 57.19\msf{1.49} & 70.97\msf{0.89} & 64.65\msf{6.73}\\
\cmidrule(l{3pt}r{3pt}){1-1} \cmidrule(l{3pt}r{3pt}){2-2} \cmidrule(l{3pt}r{3pt}){3-3} \cmidrule(l{3pt}r{3pt}){4-4}
\cmidrule(l{3pt}r{3pt}){5-5} \cmidrule(l{3pt}r{3pt}){6-6} \cmidrule(l{3pt}r{3pt}){7-7} \cmidrule(l{3pt}r{3pt}){8-8}
\cmidrule(l{3pt}r{3pt}){9-9} \cmidrule(l{3pt}r{3pt}){10-10} \cmidrule(l{3pt}r{3pt}){11-11}
MixMatch-ft~\cite{berthelot2019mixmatch}&-& 43.24\msf{0.65} & 43.68\msf{3.01} & 56.79\msf{1.89} & 71.04\msf{2.28} & 57.70\msf{0.56} & 52.53\msf{0.56} & 52.78\msf{0.84} & 62.36\msf{2.10} & 60.06\msf{0.76} \\
ReMixMatch-ft~\cite{berthelot2020remixmatch}&-& 47.47\msf{1.47} & 54.39\msf{0.78} & 66.88\msf{0.38} & 78.95\msf{0.67} & 62.30\msf{1.32} & 55.48\msf{0.76} & 56.63\msf{1.81} & 68.67\msf{1.24} & 65.58\msf{0.86} \\
\cmidrule(l{3pt}r{3pt}){1-1} \cmidrule(l{3pt}r{3pt}){2-2} \cmidrule(l{3pt}r{3pt}){3-3} \cmidrule(l{3pt}r{3pt}){4-4}
\cmidrule(l{3pt}r{3pt}){5-5} \cmidrule(l{3pt}r{3pt}){6-6} \cmidrule(l{3pt}r{3pt}){7-7} \cmidrule(l{3pt}r{3pt}){8-8}
\cmidrule(l{3pt}r{3pt}){9-9} \cmidrule(l{3pt}r{3pt}){10-10} \cmidrule(l{3pt}r{3pt}){11-11}
FixMatch-ft~\cite{sohn2020fixmatch}&-& 49.69\msf{0.86} & 54.35\msf{0.64} & 67.43\msf{1.37} & 78.73\msf{1.21} & 62.53\msf{2.02} & 54.84\msf{0.52} & 57.70\msf{1.62} & 69.79\msf{0.93} & 64.12\msf{1.48} \\
 \textbf{+ OpenCoS (ours)} &\checkmark&  \textbf{50.54\msf{1.19}} &  \textbf{56.89\msf{0.73}} &  \textbf{71.70\msf{0.43}} &  \textbf{81.68\msf{0.39}} & \textbf{65.73\msf{2.18}} &  \textbf{60.46\msf{1.70}} &  \textbf{60.89\msf{2.84}} &  \textbf{73.23\msf{2.12}} &  \textbf{67.33\msf{1.41}}\\
\bottomrule[1.2pt]
\end{tabular}
}
\end{adjustbox}
\vspace{0.1in}
\caption{Comparison of median test accuracy on 9 super-classes of ImageNet, 
which are obtained by grouping semantically similar classes in ImageNet; 
\emph{Dog}, \emph{Reptile}, \emph{Produce}, \emph{Bird}, \emph{Insect}, \emph{Food}, \emph{Primate}, \emph{Aquatic animal}, and \emph{Scenery}. 
The open-set unlabeled samples are from the entire ImageNet dataset of 1,000 classes.
All the benchmarks have 25 labels per class, and we report the mean and standard deviation over three runs with different random seeds and splits.
The best scores are indicated in bold. 
We denote methods handling unlabeled out-of-class samples (\ie open-set) as ``Open-SSL''.
}
\label{tbl:large}
\end{table*}

\subsection{Experiments on CIFAR datasets}\label{sec:varyingdata}
In this section, we evaluate our method on several benchmarks where CIFAR datasets are assumed to be in-class: more specifically, we consider scenarios that either CIFAR-10 or CIFAR-100 is an in-class dataset, with an out-of-class dataset of either SVHN or TinyImageNet. Additionally, we also consider a separate benchmark called \emph{CIFAR-Animals + CIFAR-Others} following the setup in the related work \cite{oliver2018realistic}: the in-class dataset consists of 6 animal classes from CIFAR-10, while the remaining samples are considered as out-of-class.
We fix every benchmark to have 50K training samples. We assume an 80\% proportion of out-of-class, \ie 10K for in-class and 40K for out-of-class samples, except for CIFAR-Animals + CIFAR-Others, which consists of 30K and 20K samples for in- and out-of-class, respectively.
We report ReMixMatch-ft + \mbox{OpenCoS} as it tends to outperform FixMatch-ft + \mbox{OpenCoS} in such CIFAR-scale experiments, while FixMatch-ft + OpenCoS does\footnote{
{Nevertheless, we observe that OpenCoS also improves the opposite choices, \ie FixMatch for CIFAR and ReMixMatch for ImageNet as presented in the supplementary materials.}
} in the large-scale ImageNet experiments in \cref{exp:imagenet}.
\cref{tbl:main} shows the results:~OpenCoS consistently improves ReMixMatch-ft, outperforming the other baselines simultaneously.
For example, OpenCoS improves the test accuracy of ReMixMatch-ft 28.51\% $\rightarrow$ 69.38\%,
also outperforming the strongest open-set SSL baseline, OpenMatch-ft of 62.71\%,
on 4 labels per class of CIFAR-10 + TinyImageNet.

Also, we observe large discrepancies between the median and best accuracy of semi-supervised learning baselines, MixMatch-ft, \mbox{ReMixMatch-ft}, and FixMatch-ft,
especially in the extreme label-scarce scenario of 4 labels per class, \ie these methods suffer from over-fitting on out-of-class samples.
One can also confirm this significant over-fitting in state-of-the-art SSL methods by comparing other baselines with detection schemes, \eg USAD-ft, RealMix-ft, DS$^3$L-ft, and OpenMatch-ft,
which show less over-fitting but with lower best accuracy.

\subsection{Experiments on ImageNet datasets}\label{exp:imagenet}

We also evaluate OpenCoS on ImageNet to verify its scalability to a larger and more complex dataset.
We design 9 benchmarks from ImageNet dataset, similarly to Restricted ImageNet~\cite{tsipras2018robustness}: 
more specifically, we define 9 super-classes of ImageNet, each of which consists of 11$\sim$118 sub-classes.
We perform our experiments on each super-class as an individual dataset.
Each of the benchmarks (a super-class) contains 25 labels per sub-class, and we use the full ImageNet as an unlabeled dataset (excluding the labeled ones).
In this experiment, we checkpoint per $2^{15}$ training samples and report the median test accuracy of the last 3 out of 10.
We present additional experimental details, \eg configuration of the dataset, in the supplementary material.
\cref{tbl:large} shows the results: %
OpenCoS still effectively improves the baselines, largely surpassing SimCLR-le and SimCLR-ft as well.
For example, OpenCoS improves the test accuracy on Bird to 81.68\% %
from FixMatch-ft of 78.73\%, %
also improving SimCLR-ft of 77.37\% %
significantly.
OpenCoS also outperforms other open-set SSL baselines, such as DS$^3$L-ft of 78.00\% and OpenMatch-ft of 78.60\% on Bird.
This shows the efficacy of OpenCoS in exploiting open-set unlabeled data from unknown (but related) classes or even unseen distribution of another dataset in the real-world.

\section{Ablation study}\label{abl:total}

We perform an ablation study to understand further how OpenCoS works. Specifically, we assess the individual effects of the components in OpenCoS and show that each of them has an orthogonal contribution to the overall improvements.
We consider CIFAR-10 + SVHN benchmark with 4 labels per class, and \cref{tbl:contributions} summarizes the results.

\begin{table}[t]
\centering 
\begin{adjustbox}{max width=0.7\linewidth}
\begin{tabular}{cccc|c}
\toprule
\multicolumn{4}{c|}{\large{OpenCoS components}} & \large{In- + Out-of-class}\\
\cmidrule(l{3pt}r{3pt}){1-4} \cmidrule(l{3pt}r{3pt}){5-5} 
\large{Detect} & \large{Aux.~loss} & \large{Aux.~BNs} & \large{Top-k~PL} & \large{CIFAR-10 + SVHN} \\\midrule
- & - & - & - & \large{22.53}\msf{2.53}  \\
\checkmark & - & - & - & \large{50.87}\msf{1.27} \\
\checkmark & \checkmark & - & - & \large{55.25}\msf{1.04} \\
\checkmark & \checkmark & \checkmark & - & \large{56.70}\msf{1.35} \\
\checkmark & \checkmark & \checkmark & \checkmark & \textbf{\large{58.02}\msf{0.83}} \\
\bottomrule
\end{tabular}
\end{adjustbox}
\vspace{0.1in}
\caption{
{
Ablation study on four components of our method:
the detection criterion (``Detect''), auxiliary loss (``Aux. loss''), auxiliary BNs (``Aux. BNs''), and \emph{top-k} pseudo-labeling (``Top-k PL'').
We report the mean and standard deviation over three runs with different random seeds and a fixed split of labeled data.}
}
\label{tbl:contributions}
\end{table}
\noindent\textbf{Detecting and soft-labeling out-of-class samples.}
We first observe that our detection method (“Detect”), 
which simply applies SSL using detected in-class unlabeled samples,
fixes the failure modes of the baseline, \ie ReMixMatch-ft, from 22.53\% to 50.87\%.
Interestingly, leveraging out-of-class samples achieve significant improvements; \eg auxiliary loss (“Aux. loss”) and auxiliary BNs (“Aux. BNs”) improves the performance from 50.87\% (of “Detect”) to 56.70\%.
We emphasize that our soft-labeling scheme can be rather viewed as a more reasonable way to label such out-of-class samples compared to existing state-of-the-art SSL methods, \eg MixMatch simply assigns its sharpened predictions.
\noindent\textbf{Pseudo-labeling confident samples.} We empirically observed that pseudo-labels of confident in-class samples are more accurate than randomly chosen samples.
From this observation, we expect leveraging pseudo-labels of such confident samples overcomes the limitation of few labeled data. 
Finally, we remark that our pseudo-labeling scheme for confident in-class samples (“Top-k PL”)
improves the performance from 
56.70\% to 58.02\%.
\noindent\textbf{Actual soft-label assignments.}
We also present some concrete examples of our soft-labeling scheme for a better understanding,
which are obtained from unlabeled samples in the CIFAR-10 + TinyImageNet benchmark.
Overall, we qualitatively observe that out-of-class samples that share some semantic features to the in-classes %
(\eg \cref{fig:ex2}) have relatively high confidence capturing such similarity, 
while returning very close to uniform otherwise
(\eg \cref{fig:ex3}).

\begin{table}[t]
\centering 
\small
\begin{adjustbox}{max width=0.94\linewidth}
\begin{tabular}{cccc}
\toprule
Super-classes & {\# In-classes} & FixMatch-ft & + OpenCoS\\
\cmidrule(l{3pt}r{3pt}){1-4}
Dog & 118 & 63.29\msf{0.97} & \textbf{65.92\msf{0.21}} \\
Reptile & 36 & 64.35\msf{1.50} & \textbf{66.76\msf{0.71}} \\
Produce & 22 & 79.18\msf{0.81} & \textbf{80.94\msf{0.62}} \\
Bird & 21 & 88.57\msf{0.17} & \textbf{90.89\msf{0.38}} \\
Insect & 20 & 75.13\msf{0.40} & \textbf{76.90\msf{1.04}} \\
Food & 19 & 66.95\msf{0.59} &\textbf{72.00\msf{0.94}} \\
Primate & 18 & 73.07\msf{0.42} & \textbf{76.37\msf{0.82}} \\
Aquatic & 13 & 76.72\msf{1.48} & \textbf{80.21\msf{0.62}} \\
Scenery & 11 & 68.36\msf{1.79} & \textbf{69.82\msf{0.31}} \\
\bottomrule
\end{tabular}
\end{adjustbox}
\vspace{0.1in}
\caption{
{
Comparison of median test accuracy on 9 super-classes of ImageNet with DINO~\cite{caron2021emerging} (instead of SimCLR), which is based on ViT~\cite{dosovitskiy2020image} architecture.
We report the mean and standard deviation over three runs with different random seeds and splits.}
}
\label{tbl:vit}
\vspace{-0.2in}
\end{table}

\vspace{0.03in}
\noindent\textbf{Other pre-training scheme.}
In order to investigate the compatibility of OpenCoS with other self-supervised training schemes for pre-training, we also evaluate OpenCoS with DINO~\cite{caron2021emerging} instead of SimCLR, which is built upon a transformer-based architecture, namely Vision Transformer (ViT)~\cite{dosovitskiy2020image}.
We consider 9 ImageNet benchmarks (see \cref{exp:imagenet}) to validate the effectiveness of OpenCoS with FixMatch-ft, 
and \cref{tbl:vit} summarizes the results.
We observe that OpenCoS still consistently improves FixMatch-ft on the ViT encoder pre-trained via DINO.
For example, OpenCoS improves the test accuracy on Food to 72.00\% from FixMatch-ft of 66.95\%.
Remarkably, the performances of ViT outperform
that of ResNet (in \cref{tbl:large}),
which implies that OpenCos can be further improved under better unsupervised representations for handling open-set unlabeled data.

\section{Discussion}
\noindent\textbf{Conclusion.}
In this paper, we propose a simple and general framework for handling novel unlabeled data, 
aiming toward a more realistic assumption for semi-supervised learning.
Our key idea is (intentionally) not to use label information, \ie by relying on \emph{unsupervised} representation, when handling novel data,
which can be naturally incorporated into semi-supervised learning with our framework: OpenCoS.
In contrast to previous approaches,
OpenCoS opens a way to further utilize those open-set data by guessing their labels appropriately, which are again obtained from unsupervised learning.
We hope our work would motivate researchers to extend this framework with a more realistic assumption, \eg noisy labels \cite{wang2018iterative,lee2019robust}, imbalanced learning \cite{liu2020selectnet}.

\vspace{0.03in}
\noindent\textbf{Limitations.}
As our method benefits from larger models due to unsupervised pre-training~\cite{chen2020simple},
the performance would be limited on small or tiny architectures.
\vspace{0.03in}
\noindent\textbf{Potential negative societal impacts.}
The open-set unlabeled dataset may contain sensitive data, \eg facial images, 
as it is hard to check all collected data by humans.
If someone tries to classify such data from web-crawled datasets,
our method could make this process more easily.
For this reason, 
it is also important to consider this privacy issue.

\bibliographystyle{splncs04}
\bibliography{reference}

\appendix
\onecolumn
\begin{center}{\bf {\LARGE Supplementary Material}}
\end{center}

\section{Training details} \label{appen: setup}

\subsection{CIFAR-scale experiments}
For the experiments reported in \cref{tbl:main}, we generally follow the training details of 
FixMatch~\cite{sohn2020fixmatch}, including optimizer, learning rate schedule, and an exponential moving average. %
Specifically, we use Nesterov SGD optimizer with momentum 0.9, a cosine learning rate decay with an initial learning rate of 0.03, and an exponential moving average with a decay of 0.999. 
The batch size is 64, which is widely adopted in semi-supervised learning (SSL) methods.
We do not use weight decay for these models, as they are fine-tuned.
We use a simple augmentation strategy, \textit{i.e.}, flip and crop, as a default. 
We use the augmentation scheme of SimCLR~\cite{chen2020simple} (\ie random crop with resize, random color distortion, and random Gaussian blur) when a SSL method requires to specify a \emph{strong} augmentation strategy, \eg for consistency regularization in the SSL literature~\cite{berthelot2020remixmatch,sohn2020fixmatch}.

We choose the number of strong augmentations $K=1$ for ReMixMatch-ft, and 
the relative size of labeled and unlabeled batch $\mu=1$ for FixMatch-ft.\footnote{
We observed that the original hyperparameters $K=8$ and $\mu = 7$ are not effective as they could be sub-optimal under our realistic SSL scenario.
For a meaningful comparison, we choose the new hyperparameters $K=1$ and $\mu = 1$, which show better performances in our setup.}
In the case of ReMixMatch-ft, we do not use the ramp-up weighting function, the pre-mixup and the rotation loss, which give a marginal difference in fine-tuning, for efficient computation.\footnote{{For a fair comparison, ReMixMatch-ft + OpenCoS shares these settings.}}
For FixMatch-ft, we scale the learning rate linearly with $\mu$, as suggested by Sohn \etal~\cite{sohn2020fixmatch}.
Following Chen \etal~\cite{chensemi}, UASD-ft computes the predictions by accumulative ensembling instead of using an exponential moving average.
In the case of DS$^3$L-ft, we adopt the configuration of a weighting function, bi-level optimization, and mean square loss proposed by Guo \etal~\cite{guo2020safe}.
OpenCoS shares all hyperparameters of the baseline SSL methods, \eg FixMatch + OpenCoS shares hyperparameters of FixMatch-ft. 
For the results of ReMixMatch (from scratch) in \cref{fig:2}, we report the median accuracy of the last 10 checkpoints out of 200 checkpoints, where each checkpoint is saved per $2^{16}$ training samples. 

\subsection{ImageNet-scale experiments}\label{appendix: imagenet}
In \cref{exp:imagenet},
we introduce 9 benchmarks from ImageNet dataset, similar to Restricted ImageNet~\cite{tsipras2018robustness}.
In detail, we group together subsets of semantically similar classes into 9 different super-classes, as shown in \cref{tbl:classes}.

\begin{table}[ht]
\begin{center}
  \begin{tabular}{ccc}
    \toprule
    \textbf{Super-class} & \phantom{x} & \textbf{Corresponding ImageNet Classes} \\
    \midrule
    ``Dog'' &&   151  to 268    \\ 
    ``Reptile'' &&   33 to 68    \\
    ``Produce'' &&   936 to 957    \\
    ``Bird'' &&   80 to 100    \\
    ``Insect'' &&   300 to 319    \\
    ``Food'' &&   928 to 935 \& 959 to 969 \\
    ``Primate'' &&   365 to 382    \\
    ``Aquatic'' &&   118 to 121 \& 389 to 397 \\
    ``Scenery'' &&   970 to 980    \\
    \bottomrule
  \end{tabular}%
\end{center}
\caption{Super-classes used in 9 benchmarks from ImageNet dataset. The class ranges are inclusive.}
\label{tbl:classes}
\end{table}
For the experiments reported in \cref{tbl:large}, 
we use an ImageNet pre-trained ResNet-50~\cite{he2016deep} model\footnote{\url{https://github.com/google-research/simclr}.
} of SimCLR~\cite{chen2020simple}.
We follow the optimization details of the fine-tuning experiments of Chen \etal~\cite{chen2020simple}:
specifically, we use Nesterov SGD optimizer with momentum 0.9, and a learning rate of 0.00625 (following LearningRate = $0.05\cdot \text{BatchSize}/256$).
We set the batch size to 32, and report the median accuracy of the last 3 checkpoints out of 10 checkpoints in total.
Data augmentation, regularization techniques, and other hyperparameters are the same as CIFAR experiments.
In the case of FixMatch-ft + OpenCoS in \cref{tbl:large},
we empirically observe that
it is more beneficial not to discard the detected out-of-class samples for the loss of FixMatch, as it 
performs better than using in-class samples only: 
our pseudo- and soft-labeling schemes still utilize the detected in- and out-of-class samples, respectively.
Since the proportions of out-of-class samples in those benchmarks are extremely high (\eg 98.9\% in Scenery benchmark), it is possibly due to a decrease in the number of training data.
For the experiments reported in \cref{tbl:vit}, we use an ImageNet pre-trained ViT-S/16~\cite{dosovitskiy2020image} model of DINO~\cite{caron2021emerging}.\footnote{\url{https://github.com/facebookresearch/dino}.} 
We tested several learning rates for fine-tuning ViT-S/16 through an array of \{0.00625, 0.00125, 0.00025, 0.00005, 0.00001\}, and observed that the small value of 0.00005 shows the most stable and reasonable performances for the baseline, FixMatch-ft, on the ImageNet super-classes benchmarks. So, we choose the learning rate of 0.00005 for both FixMatch-ft and FixMatch-ft + OpenCoS, and follow the same hyperparameters for other details as the above.

\section{Ablation study on pre-training and model architectures}\label{supp:cifar}

\subsection{Evaluations of training from scratch}
We pre-train all the baselines via SimCLR for a fair comparison, as mentioned in \cref{sec:exp}.
We empirically observe that such fine-tuning strategy shows comparable or better performance compared to training from scratch:
in \cref{tbl:scratch}, we compare fine-tuned baseline methods with those trained from scratch on the CIFAR-100 + TinyImageNet benchmark assuming 
80\% proportion of out-of-class, \ie 10K samples for in-class and 40K samples for out-of-class.
For the training from scratch, we follow the training details those
originally used in each baseline method, and report the median accuracy of the last 10 checkpoints out of 500 checkpoints in total.
For example, ReMixMatch from scratch uses Adam optimizer with a fixed learning rate of 0.002, and weight decay of 0.02. 
We use the simple strategy (\textit{i.e.}, flip and crop) and RandAugment~\cite{cubuk2019randaugment} as weak and strong augmentation, respectively.
We also use the ramp-up weighting function, the pre-mixup and the rotation loss for ReMixMatch.

\begin{table*}[ht]
\centering
\begin{adjustbox}{max width=\textwidth}
\begin{tabular}{lP{1.5cm}P{1.5cm}P{1.5cm}P{1.5cm}P{1.5cm}}
\toprule
\multicolumn{2}{l}{In-class + Out-of-class} & \multicolumn{4}{c}{CIFAR-100 + TinyImageNet}  \\
\cmidrule(l{3pt}r{3pt}){1-1} \cmidrule(l{3pt}r{3pt}){2-6}
\multicolumn{2}{l}{Model architecture} & \multicolumn{2}{c}{Wide-ResNet-28-2} & \multicolumn{2}{c}{ResNet-50}  \\
\cmidrule(l{3pt}r{3pt}){1-2} \cmidrule(l{3pt}r{3pt}){3-4} \cmidrule(l{3pt}r{3pt}){5-6}
Labels per class & Open-SSL& 4 & 25 & 4 & 25 \\ 
\cmidrule(l{3pt}r{3pt}){1-1} \cmidrule(l{3pt}r{3pt}){2-2} \cmidrule(l{3pt}r{3pt}){3-3} \cmidrule(l{3pt}r{3pt}){4-4} \cmidrule(l{3pt}r{3pt}){5-5} \cmidrule(l{3pt}r{3pt}){6-6}
SimCLR-le                 &-& 18.14 & 33.48 & 28.17 & 43.76 \\
SimCLR-ft                 &-& 17.55 & 36.94 & 29.34 & 49.35 \\
SimCLR-sd                 &-& 18.43 & 35.79 & 24.71 & 42.74 \\
\cmidrule(l{3pt}r{3pt}){1-1} \cmidrule(l{3pt}r{3pt}){2-2} \cmidrule(l{3pt}r{3pt}){3-3} \cmidrule(l{3pt}r{3pt}){4-4} \cmidrule(l{3pt}r{3pt}){5-5} \cmidrule(l{3pt}r{3pt}){6-6}
UASD                      &\checkmark& \phantom{x}8.76 & 27.62 & \phantom{x}7.80 & 19.21 \\
UASD-ft                   &\checkmark& 11.48 & 27.65 & 24.93 & 45.10 \\
\cmidrule(l{3pt}r{3pt}){1-1} \cmidrule(l{3pt}r{3pt}){2-2} \cmidrule(l{3pt}r{3pt}){3-3} \cmidrule(l{3pt}r{3pt}){4-4} \cmidrule(l{3pt}r{3pt}){5-5} \cmidrule(l{3pt}r{3pt}){6-6}
RealMix                   &\checkmark& 13.15 & 36.97 & 12.15 & 28.46 \\
RealMix-ft                &\checkmark& 13.56 & 33.31 & 22.11 & 49.11 \\
\cmidrule(l{3pt}r{3pt}){1-1} \cmidrule(l{3pt}r{3pt}){2-2} \cmidrule(l{3pt}r{3pt}){3-3} \cmidrule(l{3pt}r{3pt}){4-4}  \cmidrule(l{3pt}r{3pt}){5-5}\cmidrule(l{3pt}r{3pt}){6-6}
DS$^3$L                   &\checkmark& 8.42 & 27.12 & 8.93 & 27.08 \\
DS$^3$L-ft                &\checkmark& 11.35 & 28.59 & 22.31 & 39.43 \\
\cmidrule(l{3pt}r{3pt}){1-1} \cmidrule(l{3pt}r{3pt}){2-2} \cmidrule(l{3pt}r{3pt}){3-3} \cmidrule(l{3pt}r{3pt}){4-4}  \cmidrule(l{3pt}r{3pt}){5-5}\cmidrule(l{3pt}r{3pt}){6-6}
OpenMatch                  &\checkmark& 1.47 & 11.81 & 15.10 & 29.42 \\
OpenMatch-ft               &\checkmark& 12.24 & 30.92 & 14.65 & 35.17 \\
\cmidrule(l{3pt}r{3pt}){1-1} \cmidrule(l{3pt}r{3pt}){2-2} \cmidrule(l{3pt}r{3pt}){3-3} \cmidrule(l{3pt}r{3pt}){4-4} \cmidrule(l{3pt}r{3pt}){5-5} \cmidrule(l{3pt}r{3pt}){6-6}
MixMatch                  &-& 14.83 & 37.94 & 14.19 & 32.49 \\
MixMatch-ft               &-& 13.11 & 39.66 & 21.58 & 51.16 \\
\cmidrule(l{3pt}r{3pt}){1-1} \cmidrule(l{3pt}r{3pt}){2-2} \cmidrule(l{3pt}r{3pt}){3-3} \cmidrule(l{3pt}r{3pt}){4-4} \cmidrule(l{3pt}r{3pt}){5-5} \cmidrule(l{3pt}r{3pt}){6-6}
FixMatch                  &-& 20.68 & 44.33 & 27.75 & 44.78 \\
FixMatch-ft               &-& 16.59 & 36.84 & 32.05 & 47.09 \\
\cmidrule(l{3pt}r{3pt}){1-1} \cmidrule(l{3pt}r{3pt}){2-2} \cmidrule(l{3pt}r{3pt}){3-3} \cmidrule(l{3pt}r{3pt}){4-4} \cmidrule(l{3pt}r{3pt}){5-5} \cmidrule(l{3pt}r{3pt}){6-6}
ReMixMatch                &-& 16.06 & 40.45 & 17.87 & 39.86 \\
ReMixMatch-ft             &-& 16.94 & 45.21 & 22.72 & 48.42 \\
\textbf{+ OpenCoS (ours)} &\checkmark& \textbf{26.51} & \textbf{47.15} & \textbf{36.38} & \textbf{55.74} \\
\bottomrule
\end{tabular}
\end{adjustbox}
\vspace{0.1in}
\caption{Comparison of the median test accuracy of Wide-ResNet-28-2 and ResNet-50 on CIFAR-100 + TinyImageNet benchmark over baseline methods. The best scores are indicated in bold. We denote methods handling unlabeled out-of-class samples (\ie open-set) as ``Open-SSL''.
We report the performance of a single run with a fixed split of labeled data.}\vspace{0.05in}
\label{tbl:scratch}
\end{table*}

\subsection{Analysis of model architectures}
For all our experiments, we use ResNet-50 following the standard of SimCLR~\cite{chen2020simple}.
This architecture is larger than Wide-ResNet-28-2~\cite{zagoruyko2016wide}, a more widely adopted architecture in the semi-supervised learning literature~\cite{oliver2018realistic}. 
We have found that using a larger network, \ie ResNet-50, is necessary to leverage the pre-trained features of SimCLR:
in \cref{tbl:scratch}, we provide an evaluation on another choice of model architecture, \ie Wide-ResNet-28-2.
The hyperparameters are the same as the experiments on ResNet-50.
Here, one can observe that OpenCoS trained on Wide-ResNet-28-2 still improves ReMixMatch-ft, outperforming the other baselines.
More importantly, however, we observe that pre-training Wide-ResNet-28-2 
via SimCLR does not significantly improve the baselines trained from scratch, contrary to the results of ResNet-50.
As also explored by \cite{chen2020simple}, we suspect this is due to that pre-training via SimCLR requires a larger model in practice, and suggest future SSL research to explore larger architectures to incorporate more rich features into their methods, \eg features learned via unsupervised learning \cite{hnaff2019cpcv2,chen2020simple,chen2020big,chen2020mocov2}.

\section{Different choices of backbone SSL algorithms for OpenCoS}\label{supp:counterpart}
In \cref{sec:varyingdata} and \cref{exp:imagenet}, we reported the performance of ReMixMatch-ft + OpenCoS and FixMatch-ft + OpenCoS for CIFAR and ImageNet datasets, respectively, as it tends to outperform the opposite choices with a large margin.
Nonetheless, we observe that our method also improves FixMatch-ft for CIFAR and ReMixMatch-ft for ImageNet experiments as shown in \cref{tbl:counterpart_cifar} and \ref{tbl:counterpart_imagenet}, respectively.
For example, OpenCoS improves the test accuracy of FixMatch-ft from 34.94\% to 73.09\% on the CIFAR-10 + TinyImageNet benchmark, and ReMixMatch-ft from 65.58\% to 67.00\% on the Scenery benchmark.

\begin{table*}[ht]
    \centering
    \footnotesize
    \begin{adjustbox}{width=0.93\textwidth}
    \begin{tabular}{lccccc}
    \toprule
        \multicolumn{1}{l}{In-class} & \multicolumn{1}{c}{CIFAR-Animals} & \multicolumn{2}{c}{CIFAR-10} & \multicolumn{2}{c}{CIFAR-100}  \\
        \cmidrule(l{3pt}r{3pt}){1-1} \cmidrule(l{3pt}r{3pt}){2-2} \cmidrule(l{3pt}r{3pt}){3-4} \cmidrule(l{3pt}r{3pt}){5-6}
        \multicolumn{1}{l}{Out-of-class} & \multicolumn{1}{c}{CIFAR-Others} & \multicolumn{1}{c}{SVHN} & \multicolumn{1}{c}{TinyImageNet} &        \multicolumn{1}{c}{SVHN} & \multicolumn{1}{c}{TinyImageNet}\\
        \midrule[0.7pt]
        \multicolumn{6}{c}{\cellcolor{gray! 20}\emph{\# labels per class = 4}}  \\
        \midrule[0.7pt]
        FixMatch-ft~\cite{sohn2020fixmatch}  & 34.94\ms{6.18} (75.83) & 32.70\ms{6.28} (55.58) & 35.99\ms{2.63} (63.35) & 23.56\ms{0.68} (24.24) & 30.70\ms{3.67} (32.52) \\
        {+ OpenCoS (ours)}  & 73.09\ms{5.96} (78.19) & 50.23\ms{2.70} (54.23)  & 55.12\ms{3.36} (59.33) &  23.57\ms{1.43} (24.31)   &     30.66\ms{0.78} (31.68)        \\
        \cmidrule(l{3pt}r{3pt}){1-1} \cmidrule(l{3pt}r{3pt}){2-2} \cmidrule(l{3pt}r{3pt}){3-4} \cmidrule(l{3pt}r{3pt}){5-6}
        ReMixMatch-ft~\cite{berthelot2020remixmatch}  & 47.61\ms{6.51} (64.06) & 24.56\ms{3.99} (47.65) & 28.51\ms{5.87} (55.68) & \phantom{ }9.36\ms{1.97} (21.30) & 22.33\ms{1.10} (29.77) \\
        \textbf{+ OpenCoS (ours)}  & \textbf{81.29\ms{0.93} (81.77)} & \textbf{66.42\ms{7.26} (66.78)}  & \textbf{69.38\ms{3.64} (70.02)} &  \textbf{30.29\ms{2.33} (30.62)}   &     \textbf{36.79\ms{0.97} (37.12)}        \\
        \bottomrule
    \end{tabular}
    \end{adjustbox}
    \vspace{0.1in}
    \caption{Comparison of median test accuracy on various CIFAR benchmarks with 4 labels per class.
    We report the mean and standard deviation over three runs with different random seeds and splits, and also report the mean of the best accuracy in parentheses.
    }\label{tbl:counterpart_cifar}
\end{table*}

\begin{table*}[ht]
\centering
\begin{adjustbox}{max width=0.93\textwidth}
{\LARGE
\begin{tabular}{lcccccccccc}
\toprule[1.2pt]
\multicolumn{1}{l}{In-class} & Dog & Reptile & Produce & Bird & Insect & Food & Primate & Aquatic  & Scenery\\
\cmidrule(l{3pt}r{3pt}){1-2} \cmidrule(l{3pt}r{3pt}){3-3} \cmidrule(l{3pt}r{3pt}){4-4} \cmidrule(l{3pt}r{3pt}){5-5} \cmidrule(l{3pt}r{3pt}){6-6} \cmidrule(l{3pt}r{3pt}){7-7} \cmidrule(l{3pt}r{3pt}){8-8} \cmidrule(l{3pt}r{3pt}){9-9}
\cmidrule(l{3pt}r{3pt}){9-9} \cmidrule(l{3pt}r{3pt}){10-10} \cmidrule(l{3pt}r{3pt}){11-11}
{Number of class} & 118 & 36 & 22 & 21 & 20 & 19 & 18 & 13 & 11\\
\cmidrule(l{3pt}r{3pt}){1-1} \cmidrule(l{3pt}r{3pt}){2-10} 
ReMixMatch-ft~\cite{berthelot2020remixmatch}& 47.47\msf{1.47} & 54.39\msf{0.78} & 66.88\msf{0.38} & 78.95\msf{0.67} & 62.30\msf{1.32} & 55.48\msf{0.76} & 56.63\msf{1.81} & 68.67\msf{1.24} & 65.58\msf{0.86} \\
{+ OpenCoS (ours)} &  47.52\msf{1.03} &  54.50\msf{0.31} &  67.33\msf{0.41} &  79.40\msf{0.61} & 63.60\msf{1.30} &  56.67\msf{0.44} &  57.55\msf{2.67} &  69.59\msf{0.76} &  67.00\msf{1.73}\\
\cmidrule(l{3pt}r{3pt}){1-1} \cmidrule(l{3pt}r{3pt}){2-2} \cmidrule(l{3pt}r{3pt}){3-3} \cmidrule(l{3pt}r{3pt}){4-4}
\cmidrule(l{3pt}r{3pt}){5-5} \cmidrule(l{3pt}r{3pt}){6-6} \cmidrule(l{3pt}r{3pt}){7-7} \cmidrule(l{3pt}r{3pt}){8-8}
\cmidrule(l{3pt}r{3pt}){9-9} \cmidrule(l{3pt}r{3pt}){10-10} 
FixMatch-ft~\cite{sohn2020fixmatch}& 49.69\msf{0.86} & 54.35\msf{0.64} & 67.43\msf{1.37} & 78.73\msf{1.21} & 62.53\msf{2.02} & 54.84\msf{0.52} & 57.70\msf{1.62} & 69.79\msf{0.93} & 64.12\msf{1.48} \\
 \textbf{+ OpenCoS (ours)} &  \textbf{50.54\msf{1.19}} &  \textbf{56.89\msf{0.73}} &  \textbf{71.70\msf{0.43}} &  \textbf{81.68\msf{0.39}} & \textbf{65.73\msf{2.18}} &  \textbf{60.46\msf{1.70}} &  \textbf{60.89\msf{2.84}} &  \textbf{73.23\msf{2.12}} &  \textbf{67.33\msf{1.41}}\\
\bottomrule[1.2pt]
\end{tabular}}
\end{adjustbox}
\vspace{0.1in}
\caption{Comparison of median test accuracy on 9 super-classes of ImageNet, 
which are obtained by grouping semantically similar classes in ImageNet.
The open-set unlabeled samples are from the entire ImageNet dataset of 1,000 classes.
All the benchmarks have 25 labels per class, and we report the mean and standard deviation over three runs with different random seeds and splits.
}
\label{tbl:counterpart_imagenet}
\end{table*}

\section{Ablation study on hyperparameters}\label{appendix: ablation}
\subsection{Effects of the temperature, loss weight, and number of confident samples}\label{appendix: hyper}
In \cref{sec:varyingdata} and \ref{exp:imagenet}, 
we perform all the experiments with the temperature $\tau=0.1$ and loss weight $\lambda=0.5$.
For the \emph{top-k} sampling of confident unlabeled samples,
we use $k=10\%$ for 4 labels per class and $k=1\%$ for 25 labels per class.
To examine the effect of those hyperparameters,
we additionally test the hyperparameters
across an array of $\tau\in\{0.1, 0.5, 1, 4\}$, $\lambda\in\{0.1, 0.5, 1, 4\}$, and $k\in\{1\%, 10\%\}$
on the CIFAR-100 + TinyImageNet benchmark with ResNet-50. 
The results are presented in \cref{tbl:hyper}.
Overall, we found our method 
achieves reasonable performance even with our naïve choice of those hyperparameters
but can be further improved by tuning them.

\begin{table*}[ht]
\centering
\begin{subfigure}{0.48\linewidth}
\centering
    \begin{adjustbox}{max width=\textwidth}
    \begin{tabular}{c|cccc} 
    \toprule
    \multicolumn{5}{c}{\cellcolor{gray! 20}\emph{$k = 1\%$}}  \\
    \midrule[0.7pt]
    \diagbox{{$\tau$}}{$\lambda$} & 0.1 & 0.5 & 1 & 4 \\ 
    \midrule
    0.1 &  37.91 & 38.20 & 36.33 & 36.55 \\
    0.5 &  37.13 & 37.99 & 37.06 & 36.30 \\
    1   &  37.02 & 36.85 & 38.01 & 38.38 \\
    4   &  38.13 & 37.57 & 37.91 & 37.86 \\
    \midrule[0.7pt]
    \multicolumn{5}{c}{\cellcolor{gray! 20}\emph{$k = 10\%$}}  \\
    \midrule[0.7pt]
    \diagbox{{$\tau$}}{$\lambda$} & 0.1 & 0.5 & 1 & 4 \\ 
    \midrule
    0.1 &  36.48 & 36.38 & 35.94 & 36.11 \\
    0.5 &  36.42 & 36.31 & 35.89 & 36.44 \\
    1   &  36.77 & 36.29 & 36.81 & 36.83 \\
    4   &  37.16 & 36.55 & 36.40 & 36.31 \\
    \bottomrule
    \end{tabular}
    \end{adjustbox}
    \caption{4 labels per class.}
\end{subfigure}
\begin{subfigure}{0.48\linewidth}
\centering
    \begin{adjustbox}{max width=\textwidth}
    \begin{tabular}{c|ccccc} 
    \toprule
    \multicolumn{5}{c}{\cellcolor{gray! 20}\emph{$k = 1\%$}}  \\
    \midrule[0.7pt]
    \diagbox{{$\tau$}}{$\lambda$} & 0.1 & 0.5 & 1 & 4 \\ 
    \midrule
    0.1 &  55.68 & 55.62 & 55.29 & 55.90 \\
    0.5 &  55.38 & 56.12 & 56.12 & 55.61 \\
    1   &  55.81 & 55.91 & 54.93 & 55.88 \\
    4   &  55.60 & 55.91 & 55.53 & 55.95 \\
    \midrule[0.7pt]
    \multicolumn{5}{c}{\cellcolor{gray! 20}\emph{$k = 10\%$}}  \\
    \midrule[0.7pt]
    \diagbox{{$\tau$}}{$\lambda$} & 0.1 & 0.5 & 1 & 4 \\ 
    \midrule
    0.1 &  55.59 & 55.74 & 55.74 & 55.53 \\
    0.5 &  55.65 & 55.78 & 55.41 & 55.21 \\
    1   &  55.90 & 55.58 & 56.21 & 55.56 \\
    4   &  56.00 & 55.75 & 56.35 & 55.98 \\
    \bottomrule
    \end{tabular}
    \end{adjustbox}
    \caption{25 labels per class.}
\end{subfigure}
\caption{{Comparison of median test accuracy on the CIFAR-100 + TinyImageNet benchmark with (a) 4 and (b) 25 labels per class, over various hyperparameters 
$\tau$, $\lambda$, and $k$.
}}\label{tbl:hyper}
\end{table*}

\subsection{Evaluations of the detection threshold}\label{abl:ths}
We additionally provide the detection performance on various proportions of out-of-class samples, \ie 40\%, 50\%, and 67\%, on CIFAR-Animal + CIFAR-Others benchmark with 4 labels per class.
For each setting, the number of out-of-class samples is fixed at 20K, while in-class samples are controlled to 20K and 10K, respectively.
We choose the same detection threshold $t:=\mu_l - 2\sigma_l$ throughout all experiments:
it is a reasonable choice, giving $\approx$ 95\% confidence if the score follows Gaussian distribution. 
\cref{supp:detect} shows the detection performance of our threshold and its applicability 
over various proportions of out-of-class samples.
Although tuning $t$ gives further improvements though (see \cref{supp:threshold}), we fix the threshold without any tuning.

\begin{table*}[ht]
\begin{subfigure}{0.45\linewidth}
\centering
    \begin{adjustbox}{max width=\textwidth}
    \begin{tabular}{lccc}
    \toprule
        Proportion of out-of-class     & 40\% & 50\% & 67\%  \\
        \cmidrule(l{3pt}r{3pt}){1-1} \cmidrule(l{3pt}r{3pt}){2-4}
        True Positive Rate (TPR)   & 63.61 & 72.19 & 81.10 \\
        True Negative Rate (TNR)   & 99.76 & 99.66 & 99.55 \\
        \cmidrule(l{3pt}r{3pt}){1-1} \cmidrule(l{3pt}r{3pt}){2-4}
        AUROC (ours)& 98.10 & 98.50 & 98.90 \\
        \bottomrule
    \end{tabular}
    \end{adjustbox}
    \caption{The detection performance of the proposed threshold $t:=\mu_l - 2\sigma_l$ on varying proportions of out-of-class.}
    \label{supp:detect}
\end{subfigure}
\begin{subfigure}{0.533\linewidth}
\centering
\begin{adjustbox}{max width=\textwidth}
\begin{tabular}{lcccc}
	\toprule
	$\eta$
	& 1 & 2 & 3 & 4 \\
	\cmidrule(l{3pt}r{3pt}){1-1} \cmidrule(l{3pt}r{3pt}){2-5}
	True Positive Rate (TPR)      %
	         & 37.48
	         & 63.61
	         & 85.55
	         & 98.62 \\
    True Negative Rate (TNR)      %
	         & 99.94
	         & 99.76
	         & 97.22
	         & 74.48\\
	\cmidrule(l{3pt}r{3pt}){1-1} \cmidrule(l{3pt}r{3pt}){2-5}
	Accuracy (ours) %
	         & 76.32
	         & 81.08
	         & 83.50
	         & 81.00 \\
	\bottomrule
\end{tabular}
\end{adjustbox}
    \caption{The detection performance and median test accuracy across different thresholds $t$, \ie $\eta=1, 2, 3, 4$ of $t=\mu_l - \eta\cdot\sigma_l$.}
    \label{supp:threshold}
\end{subfigure}
\caption{The detection performance across different (a) proportions of out-of-class and (b) detection thresholds in CIFAR-Animals + CIFAR-Others benchmark with 4 labels per class.}
\end{table*}

\section{Ablation study on out-of-class samples}

\subsection{Effects of soft-labeling}\label{appendix:soft-assign}
We again emphasize that our soft-labeling scheme can be viewed in a more reasonable way to label such out-of-class samples compared to existing state-of-the-art SSL methods.
On the other hand, a prior work~\cite{li2017learning} has a similar observation to our approach: assigning soft-labels of novel data could be beneficial for transfer learning.
This motivates us to consider an experiment to support further the claim that our soft-labeling gives informative signals:
we train a classifier 
by minimizing only the cross-entropy loss with soft-labels (\ie without in-class samples) from \textit{scratch}.
Specifically, we train a classifier from scratch only with the unlabeled out-of-class samples and their soft-labels on the CIFAR-10 benchmarks with 4 labels per class.
Although the trained classifier uses none of the in-class samples, it performs much better than (random) guessing, even close to some baselines (in \cref{tbl:ooc_only});
this supports that generated soft-labels contain informative features of in-classes.

\vspace{0.03in}
\noindent\textbf{Training details. }
We use ResNet-50 and SGD with momentum 0.9, weight decay 0.0001, and an initial learning rate of 0.1. The learning rate is divided by 10 after epochs 100 and 150, 
and total epochs are 200. We set batch size as 128, and use a simple data augmentation strategy, \ie flip and crop. %
We minimize the cross-entropy loss between the soft-labels $q(\cdot)$ and 
model predictions $f(\cdot)$, \ie $\mathcal{L}=\mathbb{H}(q(x_u^{\text{out}}), f(x_u^{\text{out}}))$.
Here, we scale the model predictions $f(x_u^{\text{out}})$ with a temperature of 4 for a stable training.

\begin{table}[ht]
\centering
\begin{tabular}{lccc} %
	\toprule
	In-class & CIFAR-Animals & \multicolumn{2}{c}{CIFAR-10} \\
	\cmidrule(l{3pt}r{3pt}){1-1} \cmidrule(l{3pt}r{3pt}){2-2} \cmidrule(l{3pt}r{3pt}){3-4}
	Out-of-class & + CIFAR-Others & + SVHN & + TinyImageNet\\
	\cmidrule(l{3pt}r{3pt}){1-1} \cmidrule(l{3pt}r{3pt}){2-2} \cmidrule(l{3pt}r{3pt}){3-3} \cmidrule(l{3pt}r{3pt}){4-4}
    \cmidrule(l{3pt}r{3pt}){1-1} \cmidrule(l{3pt}r{3pt}){2-2} \cmidrule(l{3pt}r{3pt}){3-3} \cmidrule(l{3pt}r{3pt}){4-4}
    SimCLR-le & 65.58\ms{3.51}
	         & 56.89\ms{3.19}
	         & 58.20\ms{0.88} \\
	ReMixMatch-ft & 47.61\ms{6.51}
	         & 24.56\ms{3.99}
	         & 28.51\ms{5.87} \\
	\cmidrule(l{3pt}r{3pt}){1-1} \cmidrule(l{3pt}r{3pt}){2-2} \cmidrule(l{3pt}r{3pt}){3-3} \cmidrule(l{3pt}r{3pt}){4-4}
    Aux. loss only & \textbf{39.76\ms{1.97}} & \textbf{25.97\ms{3.92}}
	         & \textbf{21.19\ms{5.49}} \\ 
	\bottomrule
\end{tabular}
\vspace{0.1in}
\caption{
Comparison of the median test accuracy of ResNet-50 trained on out-of-class samples and their soft-labels of the CIFAR-10 benchmarks with 4 labels per class. We denote the new setting of 
minimizing with the auxiliary loss as ``Aux. loss only''.
We report the mean and standard deviation over three runs with different random seeds and splits of labeled data.}\label{tbl:ooc_only}
\end{table}

\subsection{Effects of out-of-class samples in contrastive Learning}
To clarify how the improvements of OpenCoS comes from out-of-class samples, we have considered additional CIFAR-scale experiments with 4 labels per class.
We newly pre-train and fine-tune SimCLR models using in-class samples only, \ie 30,000 for CIFAR-Animals, 10,000 for CIFAR-10, CIFAR-100 benchmarks,
{and compare} 
two baselines: SimCLR-le and ReMixMatch-ft.
Interestingly, we found that just merging out-of-class samples to the training dataset
improves the {performance of SimCLR models} in several cases (see \cref{tbl:simclr_none}), \eg SimCLR-le of CIFAR-10 enhances from 55.27\% to 58.20\% with TinyImageNet.
Also, OpenCoS significantly outperforms overall baselines,
even when out-of-class samples hurt the performance of SimCLR-le or ReMixMatch-ft.
We confirm that the proposed method effectively utilizes contrastive representations of out-of-class samples beneficially, 
{compared to other SSL baselines.}
\begin{table}[ht]
\centering
\resizebox{0.95\textwidth}{!}{
\begin{tabular}{lcccccc}
	\toprule
	\multicolumn{2}{c}{In-class + Out-of-class} & CIFAR-Animals & \multicolumn{2}{c}{CIFAR-10} & \multicolumn{2}{c}{CIFAR-100} \\
	\cmidrule(l{3pt}r{3pt}){1-2} \cmidrule(l{3pt}r{3pt}){3-3} \cmidrule(l{3pt}r{3pt}){4-5} \cmidrule(l{3pt}r{3pt}){6-7}
	Method & {w/ out-of-class} & + CIFAR-Others & + SVHN & + TinyImageNet & + SVHN & + TinyImageNet \\
	\cmidrule(l{3pt}r{3pt}){1-1} \cmidrule(l{3pt}r{3pt}){2-2} \cmidrule(l{3pt}r{3pt}){3-7}
    SimCLR-le & - & 64.30\ms{6.50}
	         & \multicolumn{2}{c}{55.27\ms{2.99}}
	         & \multicolumn{2}{c}{25.17\ms{1.08}} \\
	SimCLR-le & \checkmark & 65.58\ms{3.51}
	         & 56.89\ms{3.19}
	         & 58.20\ms{0.88}
	         & 22.86\ms{0.17}
	         & 27.93\ms{0.67} \\\midrule
    ReMixMatch-ft & - & 54.04\ms{7.05}
	         & \multicolumn{2}{c}{28.15\ms{0.72}}
	         & \multicolumn{2}{c}{16.55\ms{3.04}} \\ 
	ReMixMatch-ft & \checkmark 
	         & 47.61\ms{6.51}
	         & 24.56\ms{3.99}
	         & 28.51\ms{5.87}
	         &  9.36\ms{1.97}
	         & 22.33\ms{1.10} \\
	\textbf{+ OpenCoS (ours)} & \checkmark 
	         & \textbf{81.29\ms{0.93}} & \textbf{66.42\ms{7.26}}  & \textbf{69.38\ms{3.64}} &  \textbf{30.29\ms{2.33}}   &     \textbf{36.79\ms{0.97}}\\
	\bottomrule
\end{tabular}
}
\vspace{0.1in}
\caption{
Comparison of the median test accuracy for the use of out-of-class samples on the CIFAR-scale benchmarks with 4 labels per class. We denote whether using out-of-class samples for training as ``w/ out-of-class''.
We report the mean and standard deviation over three runs with different random seeds and splits. The best scores are indicated in bold.}\label{tbl:simclr_none} 
\end{table}

\section{Comparison with the state-of-the-art method}\label{supp:openmatch_setup}
Concurrent to our work, OpenMatch~\cite{saito2021openmatch} achieved state-of-the-art performance on the CIFAR-Animals + CIFAR-Others benchmark with 50 labels per class, which assumes more labels than our setup.
To further demonstrate the effectiveness of OpenCoS, we 
evaluate the performances of ReMixMatch-ft + OpenCoS and FixMatch-ft + OpenCoS on the CIFAR-Animals benchmark with 50 labels per class.
As shown in \cref{tbl:setup_openmatch}, 
both ReMixMatch-ft + OpenCoS and FixMatch-ft + OpenCoS consistently outperform OpenMatch in this setup;
specifically, ReMixMatch-ft + OpenCoS achieves the best performance of 92.1\%, compare to OpenMatch of 89.6\%.
These results verify the effectiveness of OpenCoS on various realistic SSL scenarios.

\begin{table*}[ht]
\centering
\resizebox{0.75\textwidth}{!}{
\begin{tabular}{lccc}
	\toprule
	& OpenMatch~\cite{saito2021openmatch} & FixMatch-ft + OpenCoS (ours) & \textbf{ReMixMatch-ft + OpenCoS (ours)} \\
	\midrule
	Accuracy & 89.6 $\pm$ {0.9}
	         & 90.8 $\pm$ {0.5}
	         & \textbf{92.1} $\pm$ \textbf{0.3} \\
	\bottomrule
\end{tabular}
}
\vspace{0.1in}
\caption{Comparison of the median test accuracy on the CIFAR-Animals + CIFAR-Others benchmark with 50 labels per class. We report the mean and standard deviation over three runs with different random seeds and splits.}\label{tbl:setup_openmatch}
\end{table*}

\end{document}